\documentclass[journal]{IEEEtran}

\usepackage{graphicx}
\usepackage{amssymb,amsmath,epsfig}
\usepackage{algorithm}
\usepackage{algorithmic}
\usepackage{booktabs}
\usepackage{mathrsfs}
\usepackage{color}
\usepackage{textcomp}
\usepackage{bbding}
\usepackage{pifont}
\usepackage{wasysym}
\usepackage{amssymb}
\usepackage{subcaption}
\usepackage{cite}
\usepackage{amsmath,amssymb,amsfonts}
\usepackage{xcolor}
\usepackage{multirow}
\usepackage{epsfig}
\usepackage[colorlinks, 
            linkcolor=red, 
            anchorcolor=blue, 
            citecolor=green
            ]{hyperref}

\begin{document}
%
\title{FADPNet: Frequency-Aware Dual-Path Network for Face Super-Resolution }
%
%
%
\author{Siyu Xu,
        Wenjie Li,
        Guangwei Gao,~\IEEEmembership{Senior Member,~IEEE,}
        Jian Yang,~\IEEEmembership{Member,~IEEE,}
        \\ Guo-Jun Qi,~\IEEEmembership{Fellow,~IEEE,}
        and Chia-Wen Lin,~\IEEEmembership{Fellow,~IEEE}
\thanks{This work was supported in part by the Foundation of the State Key Laboratory of Integrated Services Networks of Xidian University under Grant No. ISN27-4, and also supported by the National Natural Science Foundation of China under Grant Nos. U24A20330 and 62361166670.~\textit{(Siyu Xu and Wenjie Li contributed equally to this work.) (Corresponding author: Guangwei Gao.)}}
\thanks{Siyu Xu, Guangwei Gao and Jian Yang are with the PCA Lab, Key Laboratory of Intelligent Perception and Systems for High-Dimensional Information of Ministry of Education, School of Computer Science and Engineering, Nanjing University of Science and Technology, Nanjing 210094, China, and also with the State Key Laboratory of Integrated Services Networks, Xidian University, Xi’an 710071, China (e-mail: xusiyu200107@163.com, \{gwgao, csjyang\}@njust.edu.cn).}
\thanks{Wenjie Li is with the School of Artificial Intelligence, Beijing University of Posts and Telecommunications, Beijing 100080, China (e-mail: lewj2408@gmail.com).}
\thanks{Guo-Jun Qi is with the Research Center for Industries of the Future and the School of Engineering, Westlake University, Hangzhou 310024, China, and also with OPPO Research, Seattle, WA 98101 USA (e-mail: guojunq@gmail.com).}
\thanks{Chia-Wen Lin is with the Department of Electrical Engineering and the Institute of Communications Engineering, National Tsing Hua University, Hsinchu 300044, Taiwan 30013, R.O.C. (e-mail: cwlin@ee.nthu.edu.tw).}
}

\markboth{IEEE Transactions on Multimedia}%
{Shell \MakeLowercase{\textit{et al.}}: Bare Demo of IEEEtran.cls for IEEE Journals}
%

\maketitle
\begin{abstract}
Face super-resolution (FSR) under limited computational budgets remains challenging. Existing methods often treat all facial pixels equally, leading to suboptimal resource allocation and degraded performance. CNNs are sensitive to high-frequency facial features such as contours and outlines, while Mamba excels at capturing low-frequency attributes like facial color and texture with lower complexity than Transformers. Motivated by this, we propose FADPNet, a Frequency-Aware Dual-Path Network that decomposes facial features into low- and high-frequency components for dedicated processing. The low-frequency branch employs a Mamba-based Low-Frequency Enhancement Block (LFEB) that integrates state-space attention with squeeze-and-excitation to restore global interactions and emphasize informative channels. The high-frequency branch uses a CNN-based Depthwise Position-aware Attention (DPA) module to refine structural details, followed by a lightweight High-Frequency Refinement (HFR) module for further frequency-specific refinement. These designs enable FADPNet to achieve a strong balance between FSR quality and efficiency, outperforming existing methods. Codes will be available at \url{https://github.com/IVIPLab/FADPNet}.
\end{abstract}


%
\IEEEpeerreviewmaketitle

\section{Introduction}
\label{sec1}

\IEEEPARstart{F}{ace} super-resolution (FSR) focuses on reconstructing high-resolution (HR) facial images from low-resolution (LR) inputs, which plays a critical role in real-world scenarios where LR faces may appear in surveillance videos, historical archives, or mobile imaging. Enhancing facial image resolution is essential for improving downstream tasks such as face recognition~\cite{schroff2015facenet} and face alignment~\cite{kim2019progressive}.

\begin{figure}[t]
\centering
\subfloat[Band-wise spectral energy ratios of intermediate features extracted by Mamba and CNN.]{
    \includegraphics[width=0.46\linewidth,trim=0 10 0 0]{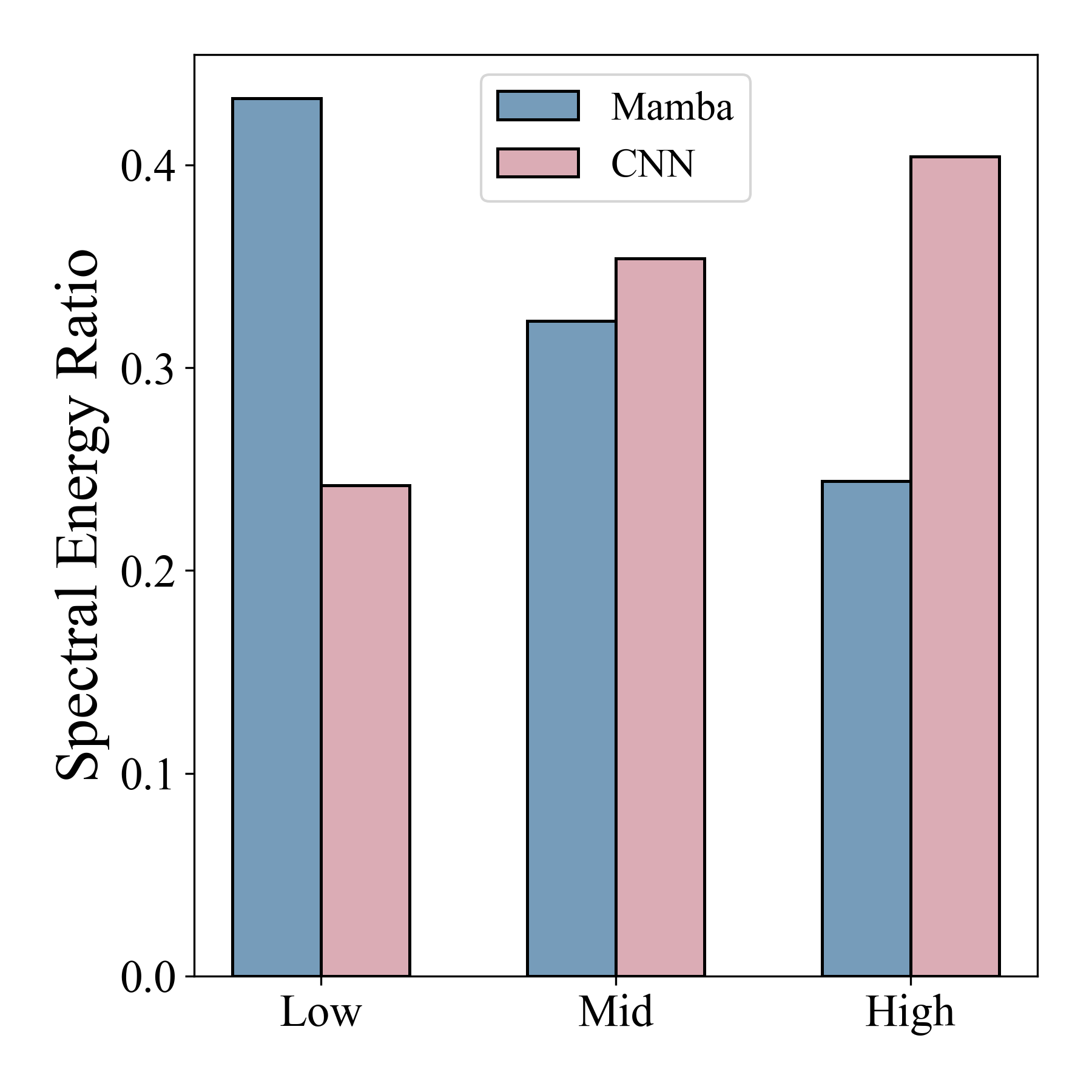}
    \label{mamba_cnn}}
\subfloat[Quality vs. Efficiency]{
    \includegraphics[width=0.45\linewidth,trim=20 10 30 30]{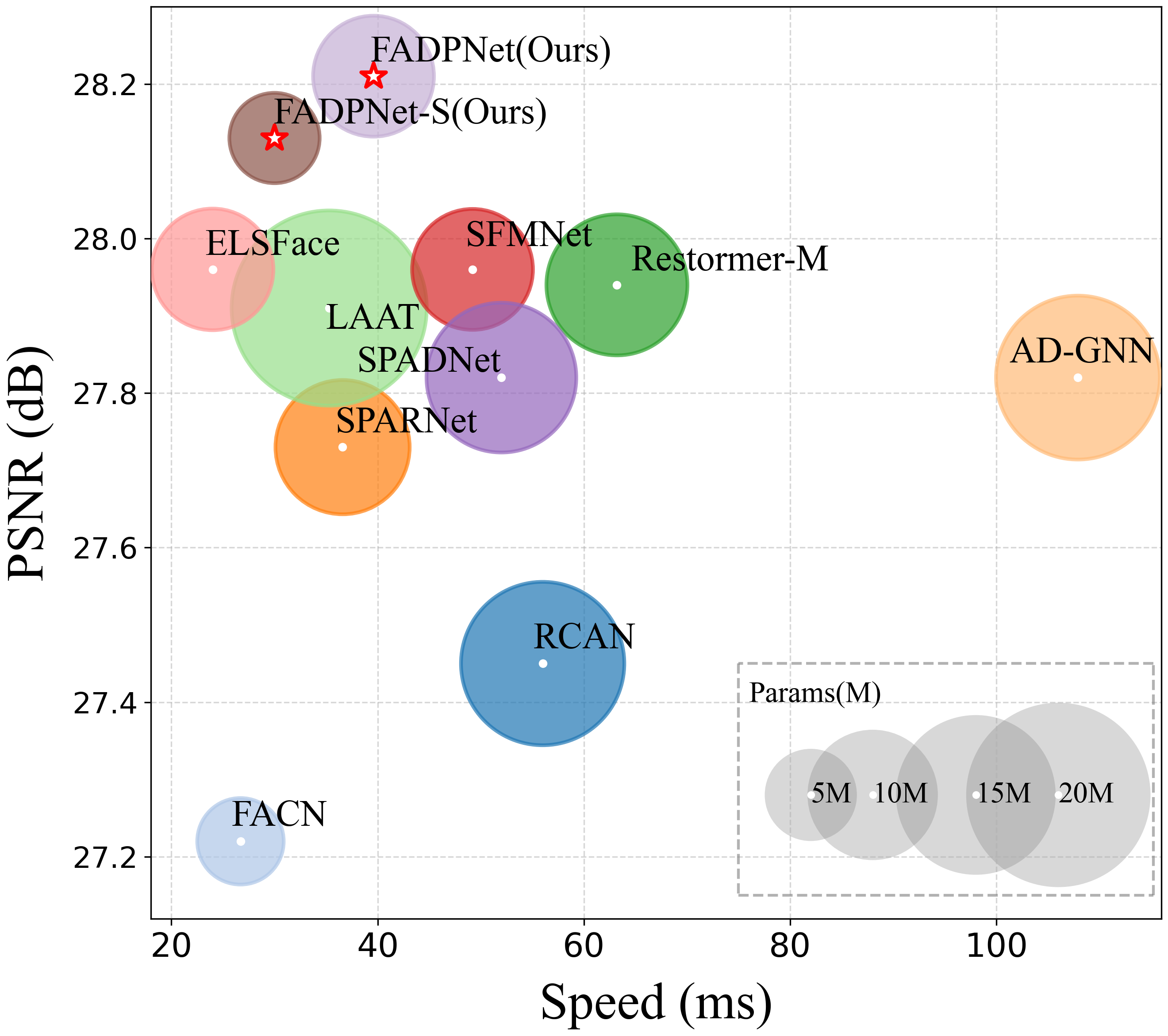}
    \label{comparison_with_models}}
\caption{(a) We divide the Fourier spectrum into low-, mid-, and high-frequency bands, and the corresponding energy ratios are averaged over the CelebA test set~\cite{liu2015deep}. Results show that Mamba preserves a noticeably larger proportion of low-frequency energy, whereas CNN retains relatively stronger high-frequency responses.
(b) Efficiency trade-offs on CelebA test set~\cite{liu2015deep}, demonstrating our method's optimal balance between PSNR, Params, and inference speed.}
\label{total}
\vspace{-3mm}
\end{figure}

One core challenge in FSR is the long-tailed distribution of image frequency components~\cite{tian2024image}. High-frequency regions, such as the eyes, nose, mouth, and facial contours, occupy only a small portion of the face yet require disproportionate modeling capacity due to their variability. They are highly sensitive to changes in identity and expression, making them more difficult~\cite{gou2023rethinking} to reconstruct than low-frequency areas like skin, which contribute little to identity-specific features. This imbalance creates a bottleneck—failure to recover high-frequency content leads to perceptual degradation. However, most existing methods treat all pixels equally, ignoring the skewed distribution of visual complexity, which results in inefficient resource use and suboptimal reconstruction. Although some works attempt frequency separation~\cite{kim2019progressive}, they often fail to balance global and local facial structures.

To address these limitations, we aim to adaptively distinguish and process high- and low-frequency information. Meanwhile, we wonder about a dual-path feature fusion to mitigate the imbalance between local and global modeling in prior frequency-aware methods. Low-frequency facial components are mainly associated with smooth, slowly varying, and spatially correlated patterns, such as coarse facial geometry, contour continuity, illumination distribution, and skin-tone consistency~\cite{zhong2018joint}. These characteristics favor state-space modeling, whose recurrent state propagation enables efficient long-range context aggregation and promotes globally coherent representations~\cite{gu2023mamba,guo2024mambair,guo2024mambairv2}. In contrast, high-frequency facial components are concentrated around localized discontinuities and rapid spatial transitions, such as eye boundaries, mouth contours, nose edges, and fine texture details~\cite{zhong2018joint,zhou2018high}. Convolutions are more effective for such content because they act on local neighborhoods and respond strongly to spatial gradients and neighborhood-level variations. Consistent with this interpretation, we compare the band-wise frequency characteristics of intermediate features extracted by CNN~\cite{ketkar2021convolutional} and Mamba~\cite{gu2023mamba}. As shown in Fig.~\ref{mamba_cnn}, Mamba preserves a noticeably larger proportion of low-frequency energy, whereas CNN retains relatively stronger high-frequency responses. These observations lead us to consider an alternative design: could we develop a two-branch architecture that allows Mamba and CNN to each specialize in the frequency domains they are most suited for, while integrating their outputs to better balance global and local facial structure representation?

Based on the above observations and analysis, we propose the Frequency-Aware Dual-Path Network (FADPNet), which allocates distinct submodules to different frequency domains. The Low-Frequency Enhancement Block (LFEB) integrates an Attentive State Space Block (ASSB) and Squeeze-and-Excitation Block (SEB) to restore stable, identity-preserving low-frequency features. The High-Frequency Enhancement Block (HFEB) combines a High-Frequency Refinement (HFR) module and Depthwise Position-aware Attention (DPA) to refine fine-grained, structure-critical details. These modules leverage complementary global and local modeling to capture facial characteristics across frequency levels. By decomposing and processing features according to frequency importance, our approach boosts facial fidelity and improves efficiency by focusing resources on perceptually critical regions. As shown in Fig.~\ref{comparison_with_models}, FADPNet achieves superior efficiency and reconstruction quality compared to existing FSR methods. 
The main contributions of this paper are summarized as follows:

\begin{itemize}
\item We propose the LFEB, which combines ASSB and SEB to enhance facial low-frequency representations by capturing global structures and emphasizing informative channel-wise responses.
\item We propose the HFEB, which combines HFR and DPA to strengthen local high-frequency facial structures and further capture long-range spatial dependencies to adaptively refine high-frequency contexts.
\item Based on the architecture designed from our frequency analysis, our FADPNet demonstrates competitive FSR ability relative to existing FSR methods across performance, model size, and inference speed.
\end{itemize}


\section{Related Work}
\label{sec2}

\subsection{Face Super-Resolution}
\label{sec21}
Face super-resolution \cite{li2025survey} has made notable strides with the advancement of deep learning. Early approaches \cite{chen2018fsrnet,kim2019progressive,ma2020deep,hu2020face} leveraged explicit facial priors to guide reconstruction. For instance, FSRNet \cite{chen2018fsrnet} incorporated facial landmarks and parsing maps to enforce structural consistency, while FAN~\cite{kim2019progressive} applied a face alignment network with progressive training to enhance realism. 
DIC~\cite{ma2020deep} introduced an iterative refinement loop combining facial components and landmark predictions. To improve the representation of facial spatial features, Hu \textit{et al.}~\cite{hu2020face} first utilized 3D shape priors in the network to better preserve sharp facial spatial structures. 

However, prior-based methods rely heavily on accurate prior estimation, which is unreliable at extremely low resolutions and adds computational overhead. To address this, recent works have explored attention-based~\cite{chen2020learning} or data-driven~\cite {peng2025towards} strategies. SPARNet~\cite{chen2020learning} used spatial attention to focus on key facial regions, SISN~\cite{lu2021face} decoupled attention into structural and textural streams, and AD-GNN~\cite{bao2022attention} combined spatial attention with graph-based message passing. LAAT~\cite{li2023learning} introduced local-aware refinement within a Transformer framework, SCTANet~\cite{bao2023sctanet} jointly modeled spatial and channel attention, SFMNet~\cite{wang2023spatial} employed parallel frequency-spatial branches for multi-scale learning, and WFEN~\cite{li2024efficient} used wavelet-based encoding to mitigate downsampling artifacts. Recent diffusion-based face restoration methods~\cite{li2025self,wang2025osdface}, such as OSDFace~\cite{wang2025osdface}, further demonstrated the growing interest in recovering high-quality facial structures and details with powerful generative priors. \emph{Nevertheless, most existing FSR methods apply uniform processing across spatial regions, ignoring the varying complexity of frequency components. In contrast, we propose a frequency-aware dual-path design that adaptively handles high- and low-frequency features at multiple scales, achieving more robust and detail-preserving reconstruction under LR conditions.}


\begin{figure*}[t]
	\centerline{\includegraphics[width=\textwidth]{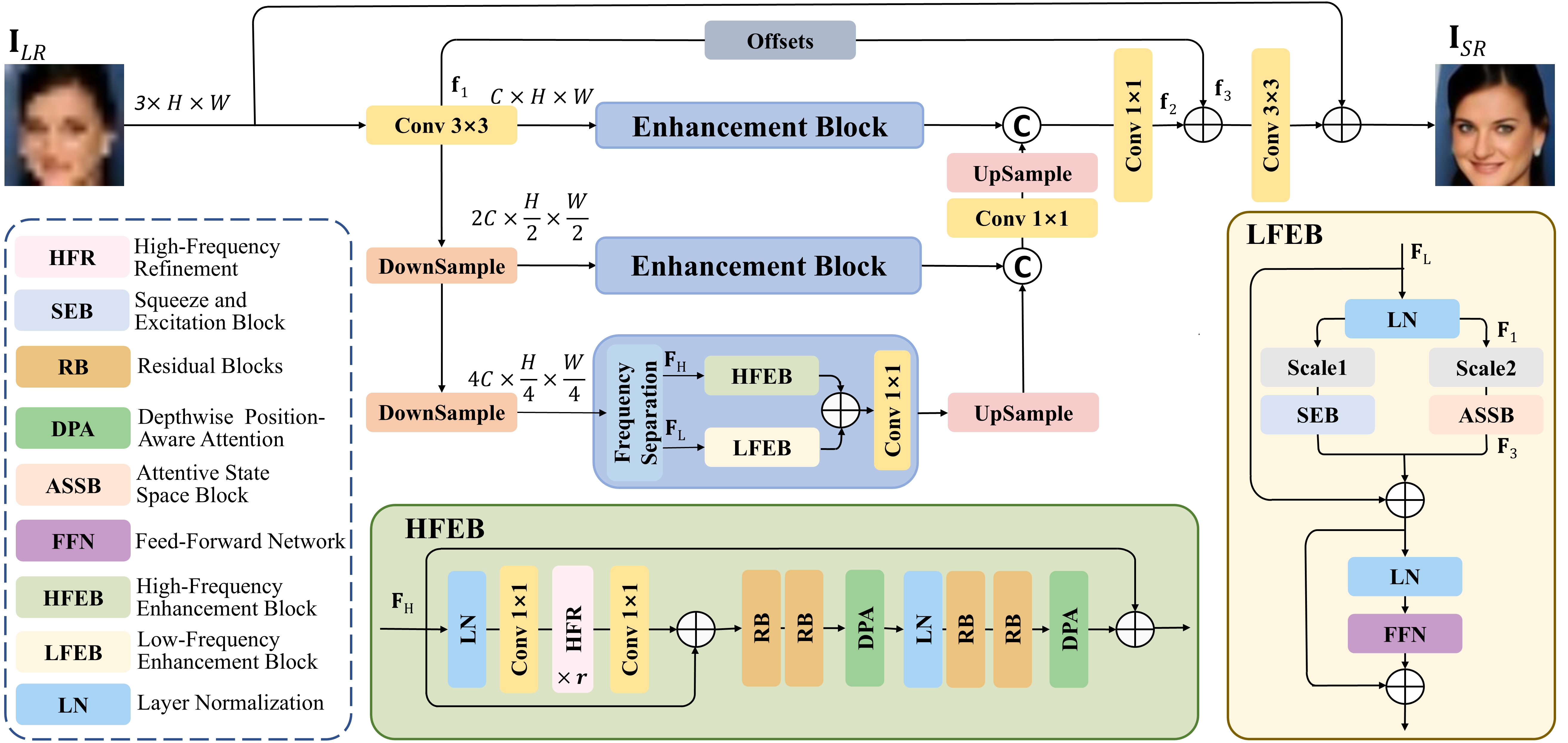}}
	\caption{Overview of our FADPNet, which adopts a U-shaped structure composed of HFEBs for high-frequency facial feature modeling and LFEBs for low-frequency facial feature enhancement. }
	\label{Network structure}
        \vspace{-3mm}
\end{figure*}

\subsection{Long-Range Modeling}
\label{sec22}
Modeling long-range dependencies is essential for FSR, as facial structures span distant spatial regions. Transformer-based methods~\cite{li2023learning,wang2023spatial} achieve strong global modeling via self-attention. For instance, LAAT~\cite{li2023learning} refined fine-grained facial structures within a Transformer framework, SFMNet~\cite{wang2023spatial} combined spatial and frequency branches to capture global and local features, and CTCNet~\cite{gao2023ctcnet} fused multi-scale global representations from Transformers.

However, vision Transformers~\cite{gao2023ctcnet,li2024efficientSR} incur quadratic complexity, limiting scalability. This has motivated exploration of lower-cost alternatives for long-range modeling. Structured State-Space Models offer linear-time global modeling. Mamba\cite{gu2023mamba} introduced a selective scan mechanism for efficient long-sequence processing with strong global context, and has been adapted to visual tasks~\cite{peng2025directing}. MambaIR~\cite{guo2024mambair} extended Mamba to image restoration, addressing local pixel forgetting and redundant channel representations, while variants like MambaLLIE~\cite{weng2024mamballie} show its adaptability to other low-level vision tasks. Despite its efficiency, Mamba inherits a causal bias from its sequence-oriented design, conflicting with the non-causal nature of super-resolution~\cite{guo2024mambair} and limiting spatial modeling. MambaIRv2~\cite{guo2024mambairv2} alleviated this via direction-agnostic scanning for richer context modeling with linear scalability. DVMSR~\cite{lei2024dvmsr} demonstrated Vision Mamba's linear-complexity global receptive field for efficient super-resolution, while MaIR~\cite{li2025mair} further validated its locality-preserving long-range modeling essential for low-frequency statistics. \emph{Unlike existing Mamba-based methods~\cite{guo2024mambair,weng2024mamballie,guo2024mambairv2}, we adopt a customized non-causal state-space backbone in the low-frequency branch, explicitly designed for FSR to better preserve structural consistency and global facial context.}

\subsection{Frequency-Aware Super-Resolution}
\label{sec23}
Frequency-aware SR methods have proven effective for enhancing structures, particularly edges and textures~\cite{cai2021freqnet,dai2024freqformer}. By decomposing image content into distinct frequency bands, they enable networks to model the characteristics of each component more precisely~\cite{li2026dual}. Low-frequency components typically represent smooth, homogeneous regions, while high-frequency components capture fine details and edges essential for perceptual quality~\cite{zhong2018joint}.

Explicit transformation methods decompose reconstruction into frequency-specific operations. Wavelet-based methods, such as SRCliqueNet~\cite{zhong2018joint}, used a discrete wavelet transform to split LR images into sub-bands, processed by specialized subnetworks to sharpen high-frequency components while preserving low-frequency structures. FreqNet~\cite{cai2021freqnet} operated in the discrete cosine transform domain, predicting HR frequency coefficients with learnable spectral decoders. In contrast, implicit methods incorporate frequency awareness into network design. SRDN~\cite{zhou2018high} employed high-pass constrained convolutions and residual connections to emphasize high-frequency propagation, 
AFD~\cite{peng2024lightweight} estimated feature drifting in the DCT domain to adaptively enhance features. Recent works such as LSRNA~\cite{jeong2025latent} further improved detail fidelity in higher-resolution diffusion-based image generation through latent-space super-resolution and region-aware enhancement. \emph{However, existing frequency-aware methods~\cite{qu2024frequency} usually improve reconstruction through frequency enhancement or decomposition, while the different modeling needs of low- and high-frequency facial information are not explicitly addressed. In contrast, our dual-branch framework separately captures global low-frequency structures and local high-frequency details, improving both reconstruction quality and efficiency.}

\section{Proposed Method}
\label{sec3}
\subsection{Overview of FADPNet}
\label{sec31}
The proposed architecture, shown in Fig.~\ref{Network structure}, follows a hierarchical frequency-aware framework. Given an input image $\mathbf{I}_{LR}\in \mathbb{R}^{3\times H\times W}$, shallow features ${\mathbf{f}_{1}}\in \mathbb{R}^{C\times H\times W}$ are first extracted. We adopt a three-level U-Net, where each stage uses a cascaded basic block to decompose features into high- and low-frequency components for separate processing. The High-Frequency Enhancement Block (HFEB) refines fine details via convolutions, Residual Blocks (RB), a High-Frequency Refinement (HFR) module, and Depthwise Position-aware Attention (DPA). The Low-Frequency Enhancement Block (LFEB) employs a dual-branch design: a Squeeze-and-Excitation Block (SEB) for local identity cues and an Attentive State Space Block (ASSB) for global structure modeling. Outputs are fused and refined through a Feed-Forward Network (FFN). To mitigate misalignment from down/up-sampling, an offset-based warping module aligns coarse structures with fine details, enhancing cross-scale consistency. Finally, multi-level features are fused and reduced to reconstruct the HR face image $\mathbf{I}_{SR}\in \mathbb{R}^{3\times H\times W}$.




\subsection{Low-Frequency Enhancement Block (LFEB)}
\label{sec33}
Our Low-Frequency Enhancement Block (LFEB) models smooth facial regions (forehead, cheeks, chin) with gradual intensity variations and limited identity-specific details. Given low-frequency features $\mathbf{F}_L\in \mathbb{R}^{C\times H\times W}$ from the frequency separation module, LFEB applies layer normalization, followed by two branches: a Squeeze-and-Excitation Block (SEB)\cite{hu2018squeeze} for adaptive channel calibration and an Attentive State Space Block (ASSB) for global structure modeling. A learnable fusion scheme with trainable scaling parameters balances the streams, after which a Feed-Forward Network (FFN)\cite{gu2023mamba} refines the output and adds it to the block input via a residual connection. This design restores coarse facial shape and alignment while preserving photorealism.

\begin{figure}[t]
\centering
\includegraphics[width=\columnwidth]{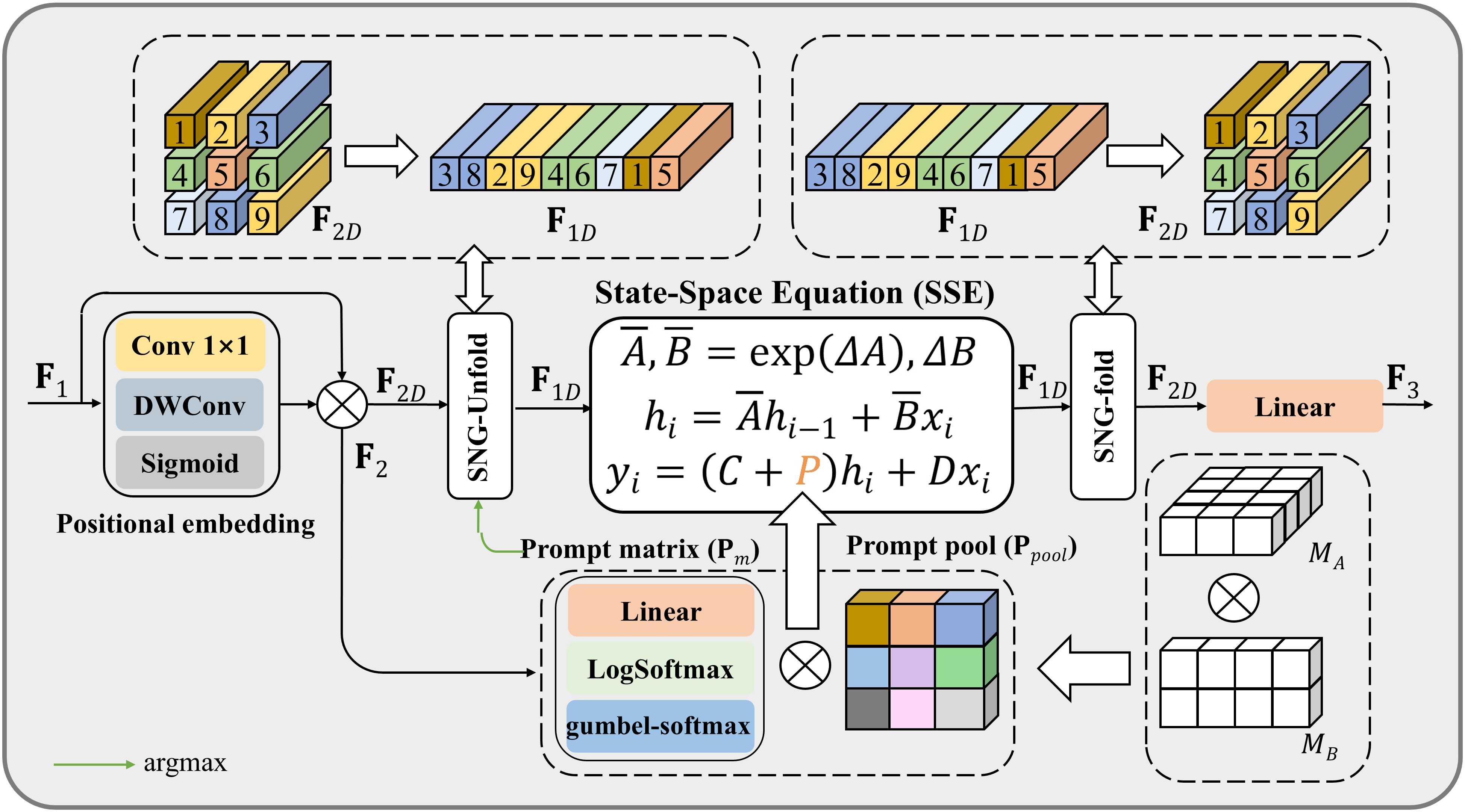}
\caption{Architecture of Attentive State Space Block (ASSB).}
\label{ASSB}
\end{figure}

\begin{figure*}[t]
\centering
\includegraphics[width=\textwidth]{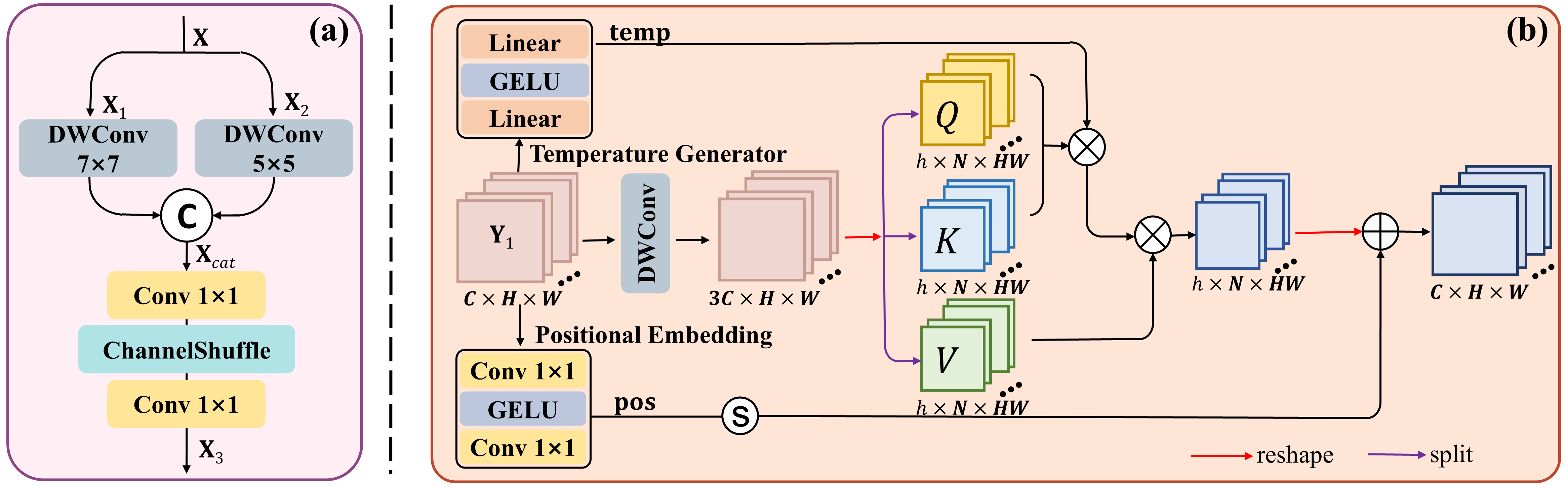}
\caption{The architecture of (a) High-Frequency Refinement (HFR), (b) Depthwise Position-aware Attention (DPA).}
\label{HEFB}
\end{figure*}

\subsubsection{Attentive State Space Block (ASSB)}
As shown in Fig.~\ref{ASSB}, building upon the Mamba~\cite{gu2023mamba} framework, ASSB introduces non-causal global modeling through an integration of semantic perceptual sequence reorganization and a cue-guided attention mechanism, which is used to address the limitations of causal modeling in the traditional state-space.  Specifically, it begins by applying a positional embedding to the input feature ${\mathbf{F}_{1}}\in \mathbb{R}^{C\times H\times W}$ to obtain a feature ${\mathbf{F}_{2}}\in \mathbb{R}^{C\times H\times W}$:
\begin{equation}
\mathbf{F}_{2} =\mathbf{F}_{1} \cdot\sigma(\mathrm{DWConv}_{3\times3}((\mathrm{Conv}_\text{1}(\mathbf{F}_{1}))),
\end{equation}
where $\mathrm{Conv}_\text{1}$ is a ${1\times1}$ convolution, $\mathrm{DWConv}_{3\times3}$ denotes a ${3\times3}$ depthwise convolution and $\sigma(\cdot)$ denotes sigmoid activation. After that, to transcend the causal constraints of standard Mamba architectures, we incorporate the learned prompts $\mathbf{P}$ into a conventional State-Space Equation (SSE), enabling richer spatial interactions beyond sequential dependency. The final prompt representation $\mathbf{P}$ consists of two components:  a prompt pool $\mathbf{P}_{\text{pool}}$ and a prompt matrix $\mathbf{P}_\text{m}$:
\begin{equation}
{{\mathbf{P}=\mathbf{P}_\text{m}\cdot \mathbf{P}_{\text{pool}}}.}
\end{equation}
The prompt pool is parameterized via low-rank decomposition for efficiency. The shared basis $\mathbf{M}_\text{A}$ captures cross-block semantic commonality, while block-specific $\mathbf{M}_\text{B}$ enables adaptive feature combination:
\begin{align}
\mathbf{P}_{\text{pool}} = \mathbf{M}_\text{A} &\cdot \mathbf{M}_\text{B}, 
\\\mathbf{M}_\text{A} \in \mathbb{R}^{T\times r},  \mathbf{M}_\text{B} &\in \mathbb{R}^{r\times d}\quad(r\ll\min\{T,d\}),
\end{align}
where ${r}$ is the inner rank, ${T}$ is the number of prompts and ${d}$ is the number of hidden states in Mamba. Instance-specific prompt matrix is governed by a routing mechanism employing $\operatorname{gumbel-softmax}$~\cite{jang2016categorical}, which generates ${L}$ instance-specific prompts ${\mathbf{P}_\text{m}\in \mathbb{R}^{L\times T}}$ dynamically associated with input pixels and relevant prompts from the pool:
\begin{equation}
\mathbf{P}_\text{m}=\operatorname{gumbel-softmax}(\operatorname{LogSoftmax}{(\mathrm{W}_\text{p}\cdot \mathbf{F}_2)),}
\end{equation}
where ${\mathrm{W}_\text{p}}$ denotes the linear layer, $\operatorname{LogSoftmax}$ refers to the logarithm of the softmax function, and gumbel-softmax is a gumbel softmax~\cite{jang2016categorical} operator. Then, the Semantic Guided Neighboring (SGN) unfolds this 2D feature $\mathbf{F}_{2}$ into 1D sequences for SSE. SGN-Unfold mechanism redefines the traditional spatial-to-sequential transformation by dynamically reorganizing pixels into 1D sequences based on semantic similarity, effectively mitigating the long-range decay inherent to existing methods~\cite{guo2024mambairv2}. Notably, the semantic information that the mechanism relies on during this process stems from the prompt matrix $\mathbf{P}_{\text{m}}$:
\begin{equation}
\mathrm{Index} = \operatorname{argmax}(\mathbf{P}_\text{m}, \text{dim}=-1).
\end{equation}
\begin{equation}
{\mathbf{F}_{1D} = \mathrm{SNG}(\mathbf{F}_{2D},\mathrm{Index})}.
\end{equation}
Tokens are subsequently sorted based on their semantic indices to group semantically similar tokens into contiguous sequences. Central to ASSB operation is SSE, a modified state-space formulation that transcends the causal constraints of standard Mamba architectures. SSE enhances the output matrix ${\mathbf{C}}$ by incorporating learnable semantic prompts—compact embeddings representing pixel groups with shared semantics. This formulation injects global contextual awareness into SSE:
\begin{equation}
{{y_i=(\mathbf{C}+\mathbf{P})h_i+\mathbf{D}x_{i}}}.
\end{equation}
Finally, an SGN-fold inversely reconstructs the semantically sorted sequence into the original 2D layout using the reverse index for precise spatial alignment. The restored feature is then linearly projected to produce the final output $\mathbf{F}_3$. Unlike conventional bidirectional Mamba variants, our single-directional scanning avoids multiple passes, reducing redundancy while maintaining global coherence. This improves inference speed and enables smooth integration with subsequent enhancement stages, where refined low-frequency features anchor high-frequency detail reconstruction.

\subsection{High-Frequency Enhancement Block (HFEB)}
\label{sec34}
In our method, the High-Frequency Enhancement Block (HFEB) targets high-frequency detail recovery. Given high-frequency features $\mathbf{F}_H\in \mathbb{R}^{C\times H\times W}$ from the frequency separation operator, we first apply a ${1\times1}$ convolution to reduce channels to $\frac{C}{2}$, then expand them back to $C$ to reduce computation. Locally, features pass through the High-Frequency Refinement (HFR) module, which progressively extracts hierarchical high-frequency patterns and strengthens details. Globally, Depthwise Position-aware Attention (DPA) modules interleaved with residual blocks selectively enhance critical high-frequency attributes while preserving realism. This combination forms the basis for high-fidelity FSR.

\subsubsection{High-Frequency Refinement (HFR)}
As shown in Fig.~\ref{HEFB} (a), our HFR module operates recurrently, to iteratively refine high-frequency components—fine-grained facial details like eyes, eyebrows, and lips—over $r$ cycles. Given an input feature $\mathbf{X}\in \mathbb{R}^{C\times H\times W}$, the module first extracts multi-scale spatial features through parallel depth-wise separable convolutions with complementary receptive fields:
\begin{equation}
\mathbf{X}_1=\mathrm{DWConv}_{7\times7}(\mathbf{X}),\quad\mathbf{X}_2=\mathrm{DWConv}_{5\times5}(\mathbf{X)},
\end{equation}
where the ${7\times7}$ kernel captures broad contextual patterns(e.g., structural contours) and the ${5\times5}$ kernel focuses on localized details(e.g., fine textures). These multi-scale outputs are concatenated along the channel dimension to obtain $\mathbf{X}_{\text{cat}}$:
\begin{equation}
\mathbf{X}_{\text{cat}}=\mathrm{Concat}[\mathbf{X}_{1},\mathbf{X}_{2}].
\end{equation}
To integrate hierarchical spatial information. Then a ${1\times1}$ convolution compresses the concatenated features into the original channel dimension while learning cross-scale correlations. To mitigate channel grouping-induced information isolation, a channel shuffle operation and a second ${1\times1}$ convolution are applied to get $\mathbf{X}_{\text{3}}$:
\begin{equation}
\mathbf{X}_{\text{3}}=\mathrm{Conv}_\text{4} \cdot \mathit{CS}(\mathrm{Conv}_\text{3}\cdot \mathbf{X}_\text{cat}),
\end{equation}
where $\mathrm{Conv}_\text{3}$ is a ${1\times1}$ convolution for channel expansion, $\mathit{CS}(\cdot)$ denotes channel-shuffle and $\mathrm{Conv}_\text{4}$ is a ${1\times1}$ convolution for channel reduction. Combined with channel shuffle, this operation dynamically permutes channel subgroups to enhance inter-channel communication. 

By fusing multi-scale context and enforcing channel diversity, the HFR amplifies suppressed high-frequency signals while filtering spatial redundancies. The refined features are then expanded back to channels and fused with residual branch outputs, providing enriched high-frequency information for subsequent components in HFEB.

\subsubsection{Depthwise Position-aware Attention (DPA)}
As shown in the Fig.~\ref{HEFB} (b), given an input feature ${\mathbf{Y}_{1}}\in \mathbb{R}^{C\times H\times W}$, a depthwise convolution ($\mathrm{DWConv}$) is first applied to ${\mathbf{f}_{in}}$, expanding its channel dimension to $3C\times H\times W$. Following established multi-head attention frameworks, we perform channel-wise partitioning into $h$ heads, each with $\frac{C}{h}$ channels. This partitioning enables concurrent learning of distinct self-attention patterns across different feature subspaces. The partitioned feature map is then rearranged into $\mathit{Q}$, $\mathit{K}$ and $\mathit{V}$ matrices of size, defined as:
\begin{equation}
{\mathit{Q,K,V}= \mathit{R S}\left(\mathrm{DWConv}_{3\times3}\left(\mathrm{Conv}_\text{5}\left(\mathbf{Y}_{1}\right)\right)\right),}
\end{equation}
where $\mathrm{Conv}_\text{5}$ is a ${1\times1}$ convolution that expands channels from $C$ to $3C$, $\mathrm{DWConv}_{3\times3}$ denotes a ${3\times3}$ depthwise convolution and $\mathit{RS}$ denotes a reshape and split operator. To enhance positional modeling capabilities, we introduce a dynamical positional encoding mechanism. The input feature map is processed through a sub-network consisting of two $1\times1\times1$ convolutions with $\operatorname{GELU}$ activation in between:
\begin{equation}
\mathbf{pos}=\mathrm{Conv}_7\cdot\operatorname{GELU}(\mathrm{Conv}_6\cdot \mathbf{Y}_{1}),\end{equation}
where $\mathrm{Conv}_6,\mathrm{Conv}_7$ denote ${1\times1}$ convolutions. However, we observe that using Positional Embedding alone did not yield the desired performance. To further refine the attention mechanism, we introduce a Temperature Generator module using a series of linear transformations to generate a temperature scale $\mathbf{temp}$, which can dynamically adjust attention scaling factors:
\begin{equation}
\mathbf{temp}=\mathrm{W}_2\cdot\operatorname{GELU}(\mathrm{W}_3\cdot \mathbf{Y}_{1}),\end{equation}
where $\mathrm{W}_{2},\mathrm{W}_{3}$ denotes the linear layer, and $\mathbf{temp}$ represents the learnable temperature parameter. Unlike fixed-temperature scaling, this adaptive mechanism provides greater flexibility in capturing long-range dependencies. In summary, the attention scores are calculated by multiplying $\mathit{Q}$ and $\mathit{K}$, and then scaled by the temperature value $\alpha$. The resulting attention-weighted $\mathit{V}$ is combined with the original positional-aware features and finally reshaped back to the original feature map size. The complete attention computation can be formulated as:
\begin{equation}
\operatorname{Attn}\mathit{\{Q,K,V \}}=\mathit{V}\cdot\operatorname{RELU}(\mathit{QK}^T\cdot\mathbf{temp})+\sigma(\mathbf{pos}),
\end{equation}
where $\sigma(\cdot)$ denotes a sigmoid activation.

\subsection{Offsets Mechanism}
\label{sec35}
As shown in Fig.~\ref{Network structure}, instead of using residual connections between input and output as in existing methods~\cite{chen2020learning,gao2023ctcnet,li2024efficient} to boost original facial information, our approach employs an offset mechanism~\cite{wang2024camixersr} to bridge the input with the residual features of multi-scale outputs. This design stems from the observation that fixed-grid sampling in traditional convolutional or upsampling operations often struggles to capture spatially variant misalignments, especially in areas with complex geometries or motion. To tackle this, our offset mechanism adaptively predicts and compensates for these misalignments by generating pixel-wise 2D displacement vectors through a lightweight convolutional predictor subnetwork, which analyzes the input features for precise feature alignment. Specifically, for the initial input $\mathbf{f}_1$:
\begin{equation}
\Delta\mathrm{~r=\mathbf{f}_\text{offsets}(\mathbf{f}_1)\in\mathbb{R}^{2\times H\times W},\mathbf{f}_3=\phi(\mathbf{f}_2,\vartriangle r),}
\end{equation}
where $\Delta\mathrm{r}$ is the content-related offsets matrix, $\phi(\cdot)$ denotes bilinear interpolation. Specifically, the offset-driven warping compensates for spatial misalignments caused by cascaded downsampling/upsampling operations, enhances high-frequency detail preservation through adaptively repositioning features according to local texture patterns, and promotes cross-scale feature consistency by aligning coarse-level structural information with fine-level details. Learned end-to-end without explicit supervision, this process allows the network to automatically find optimal spatial transformations that maximize feature coherence during reconstruction, which is especially crucial for keeping facial edge sharpness and structural integrity in FSR.

\subsection{Loss Function}
\label{Loss}
Given a training pair $\left\{\boldsymbol{I}_{LR}, \boldsymbol{I}_{HR}\right\}$, our model predicts the super-resolved image $\boldsymbol{I}_{SR}$ from the low-resolution input $\boldsymbol{I}_{LR}$. The overall training objective is defined as
\begin{equation}
\mathcal{L}_{total} = \mathcal{L}_{pixel} = \left\| \boldsymbol{I}_{SR} - \boldsymbol{I}_{HR} \right\|_1,
\end{equation}
where $\lambda=1$, and $\boldsymbol{I}_{HR}$ denotes the corresponding high-resolution ground truth. No additional perceptual, adversarial, or frequency-domain loss is employed in training.

\section{Experiments}
\label{sec4}
\subsection{Datasets and Evaluation Metrics}
\label{sec41}
We employ the CelebA~\cite{liu2015deep} dataset for training and evaluate our model on three benchmarks: CelebA~\cite{liu2015deep}, Helen~\cite{le2012interactive}, and SCface~\cite{grgic2011scface}. Therefore, we crop the image based on its center point and resize it to ${128\times128}$ pixels, serving as the high-resolution (HR) ground truth. Corresponding low-resolution (LR) inputs (${16\times16}$ pixels) are generated by downsampling HR images via bicubic interpolation. The training set comprises 18,000 images from CelebA, with 1,000 and 200 images reserved for testing and validation, respectively. For cross-dataset evaluation, we test on 50 samples from Helen and the SCface dataset without fine-tuning. We assess reconstruction quality using four widely adopted metrics: PSNR and SSIM~\cite{wang2004image} for pixel-level fidelity, LPIPS~\cite{zhang2018unreasonable} for perceptual similarity, and VIF~\cite{sheikh2006image} for texture preservation. For runtime evaluation, the reported runtime is computed as the average inference time over 1,000 test images with batch size 1. Inference is performed in FP32 precision without warmup runs, and the reported timing only includes network forward inference, excluding preprocessing and data transfer.

\subsection{Implementation Details}
\label{sec42}
Our FADPNet is implemented in PyTorch and trained on an $NVIDIA\quad RTX\quad3090\quad$ GPU. 
We initialize the model with 32 channels and design its architecture across three distinct resolution levels. Specifically, the number of low-frequency modules is 2 at each level. Meanwhile, the number of HFEB modules is configured as follows: 2 in $\mathbb{R}^{\mathrm{C}\times\mathrm{H}\times\mathrm{W}}$ stage, 3 in $\mathbb{R}^{2\mathrm{C}\times\frac{\mathrm{H}}{2}\times\frac{\mathrm{W}}{2}}$ stage, and 4 in $\mathbb{R}^{4\mathrm{C}\times\frac{\mathrm{H}}{4}\times\frac{\mathrm{W}}{4}}$ stage. Optimization is performed using the Adam optimizer (${\beta _{1} = 0.9}$ and ${\beta _{2} = 0.99}$) with an initial learning rate of ${2\times 10^{-4}}$. The model is optimized using the reconstruction loss defined in Section~\ref{Loss}, and training converges within 150 epochs using a batch size of 16. During training, standard data augmentation is applied, including random horizontal flipping and random scaling with a factor sampled from $[1.0, 1.3]$. The low-resolution input is generated by bicubic downsampling of the HR image, followed by a bicubic upsampling to the target spatial size.


\subsection{Ablation Studies}
\label{sec43}
\subsubsection{Effectiveness of HFR} To further assess the impact of the HFR module in our model, we conduct an ablation study on the CelebA test set, as summarized in Table~\ref{HFR_ablation}. First, we evaluate the model without the HFR module. Removing HFR leads to a noticeable degradation in both PSNR and SSIM, indicating that the HFR module plays a crucial role in enhancing high-frequency details. We then investigate the influence of different depth-wise convolution kernel configurations. When using only the ${5\times5}$ kernel, the model achieves a PSNR of 28.15 and SSIM of 0.8058, slightly below the performance of the configuration that employs both kernels. This suggests that the ${5\times5}$ kernel is effective at capturing localized details but insufficient for modeling a wide range of high-frequency information. In contrast, using only the ${7\times7}$ kernel yields a PSNR of 28.18 and SSIM of 0.8077, indicating its superiority in capturing broader contextual features. However, it still underperforms compared to the full HFR setup. Adding an extra ${3\times3}$ kernel results in a marginal performance drop (PSNR: 28.16, SSIM: 0.8070), implying that this addition does not significantly enhance detail restoration. Moreover, removing the channel shuffle (CS) operation causes a slight decrease in both PSNR and SSIM, highlighting the importance of inter-channel interaction in feature representation. To further validate these observations, we present residual heatmaps visualizing the pixel-wise differences between the reconstructed outputs and their HR counterparts.

\begin{figure}[t]
    \centering
	\includegraphics[width=1\columnwidth]{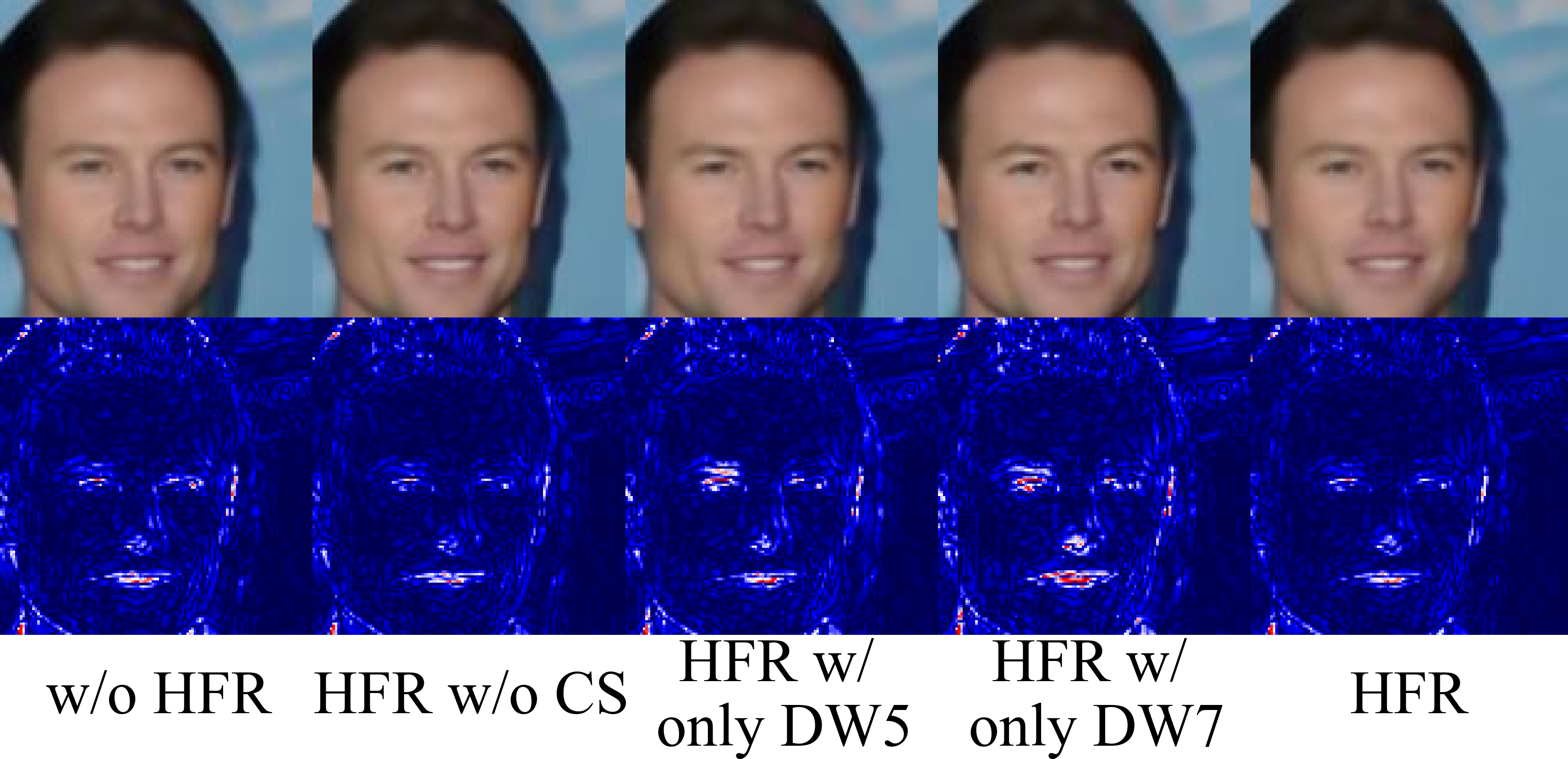}
	\caption{Comparisons of error maps on our HFR. Redder regions indicate larger pixel-wise errors. The complete HFR module in our FADPNet leads to the lowest residual errors and shows superior ability to restore high-frequency facial details.}
	\label{residual_heatmap}
\end{figure}

As shown in Fig.~\ref{residual_heatmap}, the model without the HFR module exhibits prominent errors, especially in regions rich in high-frequency textures such as facial contours. The model with HFR but without CS shows moderate improvement, suggesting that the main structure of HFR still contributes positively even without CS. Models employing only a single kind of depth-wise convolution kernel exhibit complementary characteristics: the ${5\times5}$ kernel better preserves localized features while missing global structures, whereas the ${7\times7}$ kernel captures larger-scale patterns but struggles with fine textures. In contrast, the full HFR achieves the lowest errors, visually affirming its superior capacity for high-frequency detail restoration and consistent with its leading PSNR and SSIM scores.

\subsubsection{Effectiveness of DPA} To comprehensively evaluate the individual and synergistic effects of components in our proposed Depthwise Position-aware Attention (DPA) module, we perform systematic ablation studies on the Helen test set, with detailed results presented in Table~\ref{DPA_ablation}. Our investigation focuses on three critical elements: the learnable Temperature Generator (T\_gen), fixed Temperature Parameter (T\_para), and Positional Embedding (P\_emb), analyzing their impacts on both reconstruction quality and model complexity.

The initial configuration with only T\_para provides a baseline level of reconstruction, but its fixed scaling restricts adaptability to spatial and contextual variations, resulting in suboptimal attention modulation for detailed facial regions. Adding P\_emb to T\_para slightly improves SSIM, suggesting better structural preservation, though PSNR shows a marginal decline. The P\_emb component enables more precise localization of structural details, particularly in semantically critical facial regions like ocular areas, nasal contours, and lip boundaries. Our FADPNet combines T\_gen with P\_emb, achieving peak performance. This configuration establishes a synergistic relationship where T\_gen dynamically modulates attention sharpness according to local content complexity, while P\_emb ensures spatial consistency in feature aggregation. Notably, implementing T\_gen without P\_emb guidance degrades performance. This reveals that without spatial constraints, the self-learned scaling factors may produce inconsistent attention distributions, particularly in homogeneous facial regions where positional cues are crucial for accurate feature matching.

These systematic experiments validate the necessity of integrating both adaptive temperature scaling and explicit spatial encoding in our DPA design. The final architecture achieves superior performance through three key mechanisms: (a) Content-aware attention sharpness adaptation via T\_gen, (b) Geometry-preserving feature aggregation through P\_emb, and (c) Computational efficiency from depthwise implementation. The results demonstrate our module's effectiveness in addressing the unique challenges of facial super-resolution, particularly in preserving identity-critical details while maintaining natural texture synthesis.

\begin{table}[t]
 \centering
 \caption{Ablation study of HFR on CelebA test set~\cite{liu2015deep}.}
 \centering
 \label{HFR_ablation}
 \setlength{\tabcolsep}{2.5mm} %
 \renewcommand\arraystretch{1} 
 \begin{tabular*}{\linewidth}{@{\extracolsep{\fill}}l|cc|cc}
  \toprule
  Methods & Params${\downarrow}$ & FLOPs${\downarrow}$ & PSNR${\uparrow}$ & SSIM${\uparrow}$ \\
 \hline
 \hline
  w/o HFR                                  & --     & --       & 28.16     & 0.8061 \\
  HFR w/o $\mathit{CS}$                    & 0.353M & 0.802G   & 28.20     & 0.8072 \\
  HFR w/ only $\mathrm{DWConv}_{5\times5}$ & 0.316M & 0.689G   & 28.15     & 0.8058 \\ 
  HFR w/ only $\mathrm{DWConv}_{7\times7}$ & 0.390M & 0.846G   & 28.18     & \textbf{0.8077} \\
  \textbf{HFR}                                      & 0.353M & 0.802G   & \textbf{28.21} & 0.8075 \\
  \bottomrule
 \end{tabular*}
\end{table}

\begin{table}[t]
  \caption{Ablation study of DPA on Helen test set~\cite{le2012interactive}. The last line denotes the strategy used in our final model.}
  \centering
  \renewcommand\arraystretch{1}
  \setlength{\tabcolsep}{2.5mm}
  \begin{tabular*}{\linewidth}{@{\extracolsep{\fill}}ccc|cc|cc}
    \toprule
    T\_gen & T\_para & P\_emb & Params${\downarrow}$ & FLOPs${\downarrow}$ & PSNR${\uparrow}$ & SSIM${\uparrow}$ \\
    \hline
    \hline
               & \checkmark &            & --      & --       & 26.99  & 0.8034 \\
               & \checkmark & \checkmark & 0.162M  & 0.302G   & 26.97  & 0.8037 \\
    \checkmark &            &            & 0.325M  & --       & 26.93  & 0.8008 \\
    \checkmark &            & \checkmark & 0.325M  & 0.302G   & \textbf{27.06}  & \textbf{0.8064} \\
    \bottomrule
  \end{tabular*}
  \label{DPA_ablation}
\end{table}

\begin{table}[t]
  \caption{Ablation study of ASSB on the Helen test set~\cite{le2012interactive} by replacing it with three representative alternatives:
  Self-Attention (SA)~\cite{zamir2022restormer}, Convolutional block (CNN), and Vision State Space Block (VSSB)~\cite{guo2024mambair}.}
  \centering
  \renewcommand\arraystretch{1}
  \setlength{\tabcolsep}{2mm}
  \begin{tabular*}{\linewidth}{@{\extracolsep{\fill}}l|cc|cccc}
    \toprule
     Methods & Params${\downarrow}$ & FLOPs${\downarrow}$ & PSNR${\uparrow}$  & SSIM${\uparrow}$  & VIF${\uparrow}$ & LPIPS${\downarrow}$  \\
    \hline
    \hline
    w/o ASSB   & -- & -- & 26.97 & 0.8039 & 0.4701 & 0.2202 \\
    SA         & 0.184M & 0.113G & 26.99 & 0.8035 & 0.4673 & 0.2231 \\
    CNN       & 0.287M & 0.387G & 27.01 & 0.8046 & 0.4691 & 0.2194 \\
    VSSB       & 0.269M & 0.163G & 27.02 & 0.8042 & 0.4697 & 0.2266 \\
    \textbf{ASSB}       & 0.203M & 0.125G & \textbf{27.06} & \textbf{0.8064} & \textbf{0.4706} & \textbf{0.2167} \\
    \bottomrule
  \end{tabular*}
  \label{ASSB_ablation}
\end{table}

\begin{table}[t]
    \small
    \centering
    \caption{Component-level ablation of ASSB on CelebA test set~\cite{liu2015deep} and Helen test set~\cite{le2012interactive}.}
    \centering
    \setlength{\tabcolsep}{1.8mm}
    \renewcommand\arraystretch{1}
        \begin{tabular}{l|c|c}
            \toprule
            \multirow{2}{*}{Methods} & CelebA & Helen \\
             & PSNR$\uparrow$ \ / \ SSIM$\uparrow$ & PSNR$\uparrow$ \ / \ SSIM$\uparrow$ \\
            \hline
            \hline
            w/o prompt pool          & 28.16 \ / \ 0.8061 & 26.99 \ / \ 0.8035 \\
            w/o routing              & 28.18 \ / \ 0.8069 & 27.00 \ / \ 0.8040 \\
            w/o semantic reorder    & 28.18 \ / \ 0.8074 & 26.99 \ / \ 0.8039 \\
            w/o low-rank prompt     & 28.19 \ / \ 0.8074 & 27.02 \ / \ 0.8045 \\
            \textbf{Full ASSB}   & \textbf{28.21 \ / \ 0.8075} & \textbf{27.06 \ / \ 0.8064} \\
            \bottomrule
        \end{tabular}
    \label{tab:assb_component}
\end{table}

\subsubsection{Effectiveness of ASSB} To evaluate the suitability of the Attentive State Space Block (ASSB) for low-frequency modeling in our method, we conduct ablation studies by replacing it with three representative alternatives: Self-Attention (SA)~\cite{zamir2022restormer}, Convolutional block (CNN), and Vision State Space Block (VSSB)~\cite{guo2024mambair}. As shown in Table~\ref{ASSB_ablation}, our full model with ASSB achieves the best overall performance, indicating superior perceptual quality and structural fidelity. 

It is worth noting that ASSB is not our original design, but a recently proposed state-space module tailored for long-range dependency modeling. In our framework, we integrate ASSB within LFEB to exploit its strength in global structure preservation. Unlike CNN, which tends to focus on local or high-frequency features, ASSB’s non-causal state-space formulation excels at capturing smooth, spatially consistent facial regions such as contours and skin tones. Its built-in low-rank prompt decomposition also aids in suppressing high-frequency noise and promotes stable propagation across homogeneous areas. Compared to the more general-purpose VSSB, ASSB offers a lighter structure and better frequency selectivity when dealing with low-frequency signals. Despite using fewer parameters (0.203M vs. 0.269M), ASSB outperforms or matches VSSB in all metrics, demonstrating its superior efficiency-to-performance ratio when applied to low-frequency pathways. These results collectively justify our choice of ASSB as the most effective backbone for global low-frequency modeling in our frequency-aware architecture. Its integration helps maintain large-scale facial coherence and mitigates structural misalignment—challenges that conventional transformer or Conv-based designs struggle to address.

We also conduct ablation studies by removing or simplifying individual components in ASSB while keeping the rest of the architecture unchanged. These components include the prompt pool, routing strategy, semantic reorder, and low-rank prompt design. Together, these ablations cover the key prompt-guided and sequence-organization elements in ASSB, where the semantic reorder variant also reflects the contribution of the non-causal scanning behavior introduced by sequence reorganization. As shown in Table~\ref{tab:assb_component}, simplifying any of these components leads to a performance drop on both datasets. This indicates that the improvement brought by ASSB does not arise from naive module stacking, but from the complementary effects of its internal components. In particular, the prompt-related design, routing strategy, semantic reorder, and low-rank prompt adaptation all contribute to the final performance. Meanwhile, the number of parameters and FLOPs remains nearly unchanged across variants, since these ablations mainly affect internal feature organization and prompt generation strategies rather than the overall backbone width or depth.

\begin{table}[t]
  \caption{Ablation study of our attention strategy on CelebA test set~\cite{liu2015deep} and Helen test set~\cite{le2012interactive}.}
  \centering
    \setlength{\tabcolsep}{1.8mm}
  \renewcommand\arraystretch{1}
  \begin{tabular}{l|c|c}
    \toprule
    \multirow{2}{*}{Methods} & CelebA & Helen\\
            & PSNR$\uparrow$ \ / \ SSIM$\uparrow$ & PSNR$\uparrow$ \ / \ SSIM$\uparrow$ \\
    \hline
    \hline
    LFEB w/o SEB                                 & 28.16 \ / \ 0.8067   & 26.99 \ / \ 0.8043 \\
    HFEB w/o DPA                                 & 28.04 \ / \ 0.8032   & 26.80 \ / \ 0.7981 \\
    w/o offsets                                  & 28.17 \ / \ 0.8073   & 27.00 \ / \ 0.8050   \\
    \textbf{Ours}                             & \textbf{28.21 \ / \ 0.8075}   & \textbf{27.06 \ / \ 0.8064} \\
    \bottomrule
  \end{tabular}
  \label{Frequency_ablation}
\end{table}

\begin{table}[t]
    \centering
    \caption{Ablation study of the offset-based alignment mechanism on CelebA test set~\cite{liu2015deep} and Helen test set~\cite{le2012interactive}.}
    \setlength{\tabcolsep}{1.8mm}
    \renewcommand\arraystretch{1}
        \begin{tabular}{l|c|c}
            \toprule
            \multirow{2}{*}{Methods} & CelebA & Helen \\
             & PSNR$\uparrow$ \ / \ SSIM$\uparrow$ & PSNR$\uparrow$ \ / \ SSIM$\uparrow$ \\
            \hline
            \hline
            w/o offset          & 28.17 \ / \ 0.8073 & 27.00 \ / \ 0.8050 \\
            w/ fixed warp        & 28.20 \ / \ 0.8068 & 27.01 \ / \ 0.8047 \\
            w/ zero-offset warp  & 28.18 \ / \ 0.8069 & 27.00 \ / \ 0.8033 \\
            w/ conv alignment    & 28.18 \ / \ 0.8065 & 27.00 \ / \ 0.8049 \\
            \textbf{Offset (Ours)} & \textbf{28.21 \ / \ 0.8075} & \textbf{27.06 \ / \ 0.8064} \\
            \bottomrule
        \end{tabular}
    \label{tab:offset_ablation}
\end{table}

\begin{figure}[t]
\centering
\subfloat[]{
    \includegraphics[width=0.40\linewidth,trim=30 5 0 0]{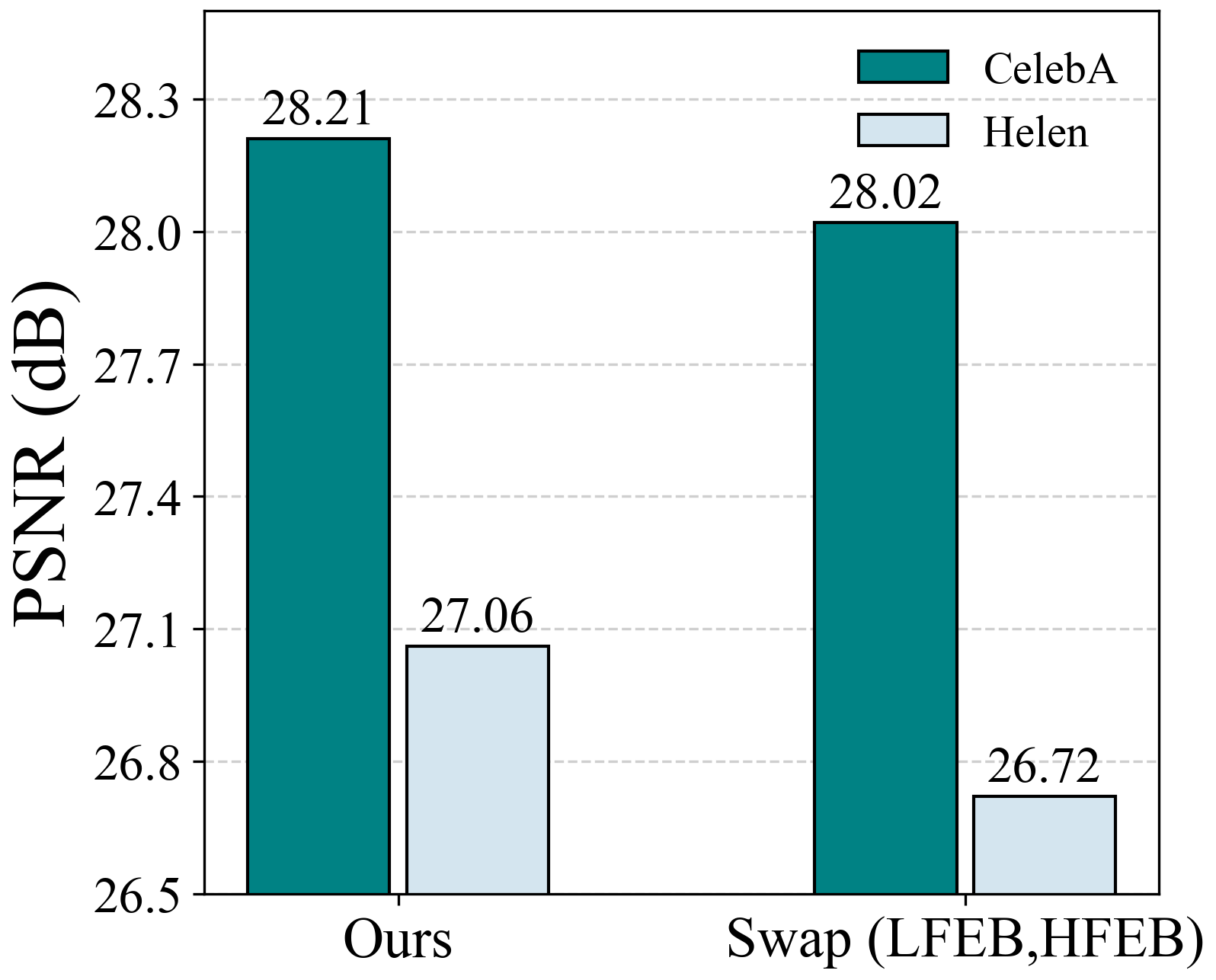}
    \label{lfeb_hfeb}}
\subfloat[]{
    \includegraphics[width=0.48\linewidth,trim=15 0 10 0]{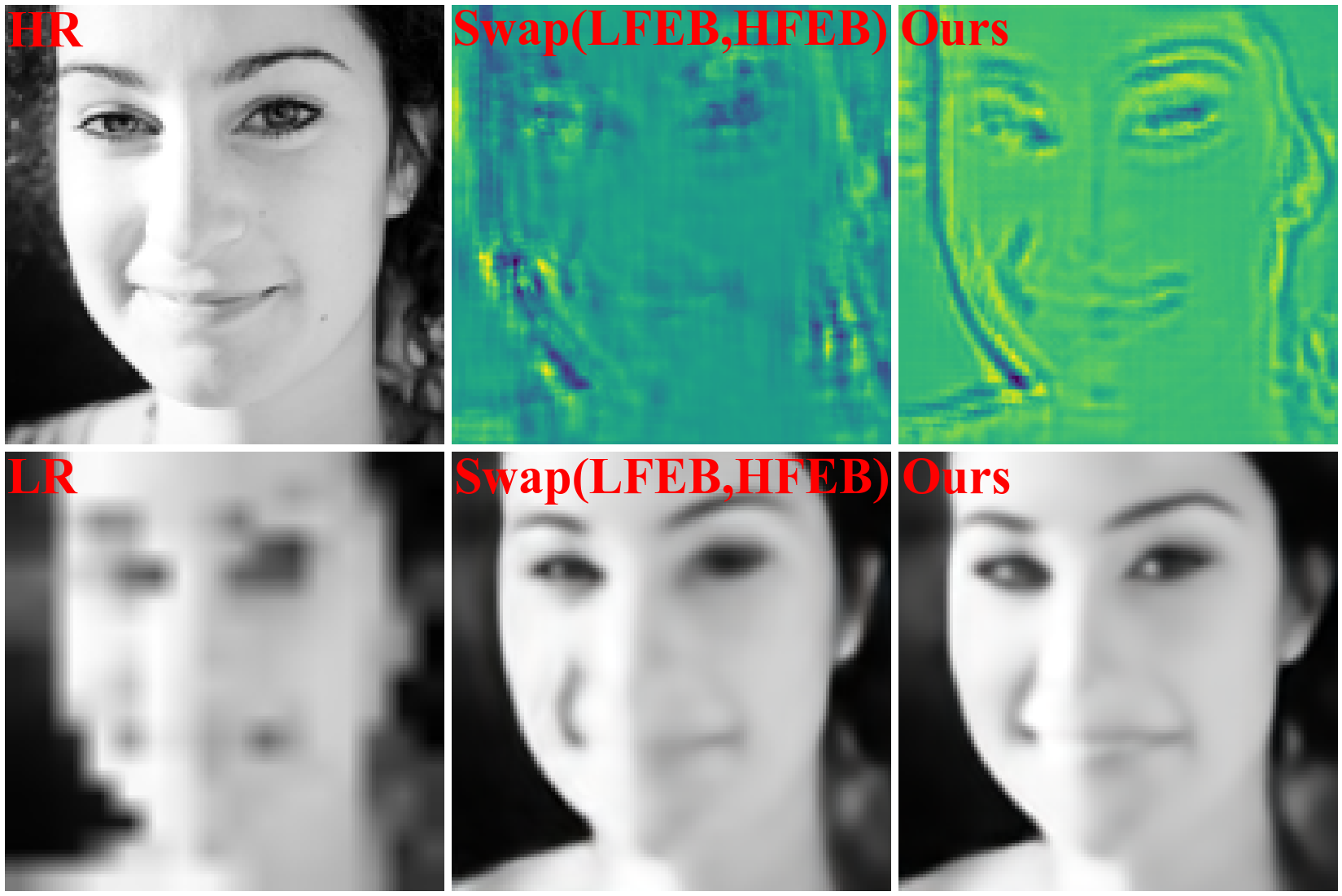}
    \label{swap}}
\caption{(a) Ablation studies of frequency-specific modules on the CelebA~\cite{liu2015deep} and Helen datasets~\cite{le2012interactive}. ``swap'' is our model in which the low- and high-frequency blocks are swapped, resulting in a noticeable performance drop compared to our original model. (b) Feature map visualization and FSR results between our original model and swapped-frequency variant show that our module design for different frequency features produces sharp and accurate facial contours.}
\label{swap ablation }
\end{figure}

\subsubsection{Study of the architecture of low- and high-frequency modules} To investigate the effectiveness of low- and high-frequency architecture, we conduct an ablation study on the CelebA and Helen datasets by interchanging the positions of LFEB and HFEB. As illustrated in Fig.~\ref{lfeb_hfeb}, this modification leads to a consistent drop in PSNR across both datasets, highlighting the critical importance of the original frequency-aware design. The original model achieves superior reconstruction performance by appropriately allocating low- and high-frequency processing to their respective branches. In contrast, the swapped configuration disrupts this balance, resulting in suboptimal representation and degraded image quality. Fig.~\ref{swap} visually compares the original and the swapped variants. The results clearly show that the proposed model preserves facial structures and fine details more effectively, yielding sharper contours and more realistic textures. Conversely, the swapped version produces noticeably blurrier outputs, especially around facial components like eyes and mouth, further validating the necessity of maintaining the original design of frequency-specific modules.

\begin{table*}[t]
    \centering
    \caption{Quantitative comparisons for $\times$8 FSR on CelebA test set~\cite{liu2015deep} and Helen test set~\cite{le2012interactive}.}
    \setlength{\tabcolsep}{2.8mm}
    \renewcommand\arraystretch{1}
    \scalebox{1}{
        \begin{tabular}{l|cc|cccc|cccc}
            \toprule
            \multirow{2}{*}{Methods} & \multirow{2}{*}{Params$\downarrow$} & \multirow{2}{*}{Speed$\downarrow$}  & \multicolumn{4}{c|}{CelebA}  & \multicolumn{4}{c}{Helen} \\
            & & & PSNR$\uparrow$ & SSIM$\uparrow$ & VIF$\uparrow$ & LPIPS$\downarrow$ & PSNR$\uparrow$ & SSIM$\uparrow$ & VIF$\uparrow$ & LPIPS$\downarrow$ \\
            \hline
            \hline
            Bicubic & -- & -- & 23.61 & 0.6779 & 0.1821 & 0.4899 & 22.95 & 0.6762 & 0.1745 & 0.4912 \\
            FSRNet~\cite{chen2018fsrnet} & 27.5M  & 89.8ms & 27.05 & 0.7714 & 0.3852 & 0.2127 & 25.45 & 0.7364 & 0.3482 & 0.3090 \\
            DIC~\cite{ma2020deep} & 22.8M  & 120.5ms & 27.42 & 0.7840 & 0.4234 & 0.2129 & 26.15 & 0.7717 & 0.4085 & \textbf{0.2158} \\
            SPARNet~\cite{chen2020learning} & 10.6M  & 36.6ms & 27.73 & 0.7949 & 0.4505 & 0.1995 & 26.43 & 0.7839 & 0.4262 & 0.2674 \\
            SISN~\cite{lu2021face}  & 9.8M  & 68.0ms & 27.91 & 0.7971 & 0.4785 & 0.2005 & 26.64 & 0.7908 & 0.4623 & 0.2571 \\
            AD-GNN~\cite{bao2022attention} & 15.8M  & 107.9ms & 27.82 & 0.7962 & 0.4470 & 0.1937 & 26.57 & 0.7786 & 0.4363 & 0.2423 \\
            Restormer-M~\cite{zamir2022restormer} & 11.7M  & 63.2ms & 27.94 & 0.8027 & 0.4624 & 0.1933 & 26.91 & 0.8013 & 0.4595 & 0.2432 \\
            LAAT~\cite{li2023learning} & 22.4M  & 35.3ms & 27.91 & 0.7794 & 0.4624 & 0.1879 & 26.89 & 0.8005 & 0.4569 & 0.2255 \\
            ELSFace~\cite{qi2023efficient} & 8.7M  & \textbf{24.0ms} & 27.96 & 0.7992 & 0.4783 & 0.1927 & 26.70 & 0.7910 & 0.4442 & 0.2495 \\ 
            SFMNet~\cite{wang2023spatial} & \underline{8.6M}  & 49.2ms & 27.96 & 0.7996 & 0.4644 & 0.1937 & 26.86 & 0.7987 & 0.4573 & 0.2322 \\ 
            SPADNet~\cite{wang2024structure} & 13.2M  & 52.0ms & 27.82 & 0.7966 & 0.4589 & 0.1987 & 26.41 & 0.7857 & 0.4295 & 0.2645 \\     
            FreMamba~\cite{xiao2024frequency} & 7.2M  & 63.0ms & 28.13 & 0.8050 & 0.4797 & 0.1807 & 26.81 & 0.7967 & 0.4542 & 0.2387 \\
            DMNet~\cite{li2026dual} & 8.3M  & 296.7ms & \underline{28.15} & 0.8045 & 0.4802 & \underline{0.1798} & 26.92 & 0.7985 & 0.4664 & 0.2335 \\
            FreqFormer~\cite{dai2024freqformer} & 9.9M  & 75ms & \underline{28.15} & 0.8047 & \textbf{0.4883} & 0.1851 & 26.82 & 0.7973  & \underline{0.4678} & 0.2290\\
             \textbf{FADPNet-S} & \textbf{4.8M}  & \underline{30.0ms} & 28.13 & \underline{0.8055} & 0.4848 & 0.1807 & \underline{26.95} & \underline{0.8015} & 0.4652 & 0.2233 \\
            \textbf{FADPNet} & \underline{8.6M}  & 39.6ms & \textbf{28.21} & \textbf{0.8075} & \underline{0.4876} & \textbf{0.1741} & \textbf{27.06} & \textbf{0.8064} & \textbf{0.4706} & \underline{0.2167} \\
            \bottomrule
        \end{tabular}
    }
    \label{compare_CelebA_Helen}
\end{table*}

\begin{figure*}[t]
 \centerline{\includegraphics[width=18cm,trim=0 10 0 0]{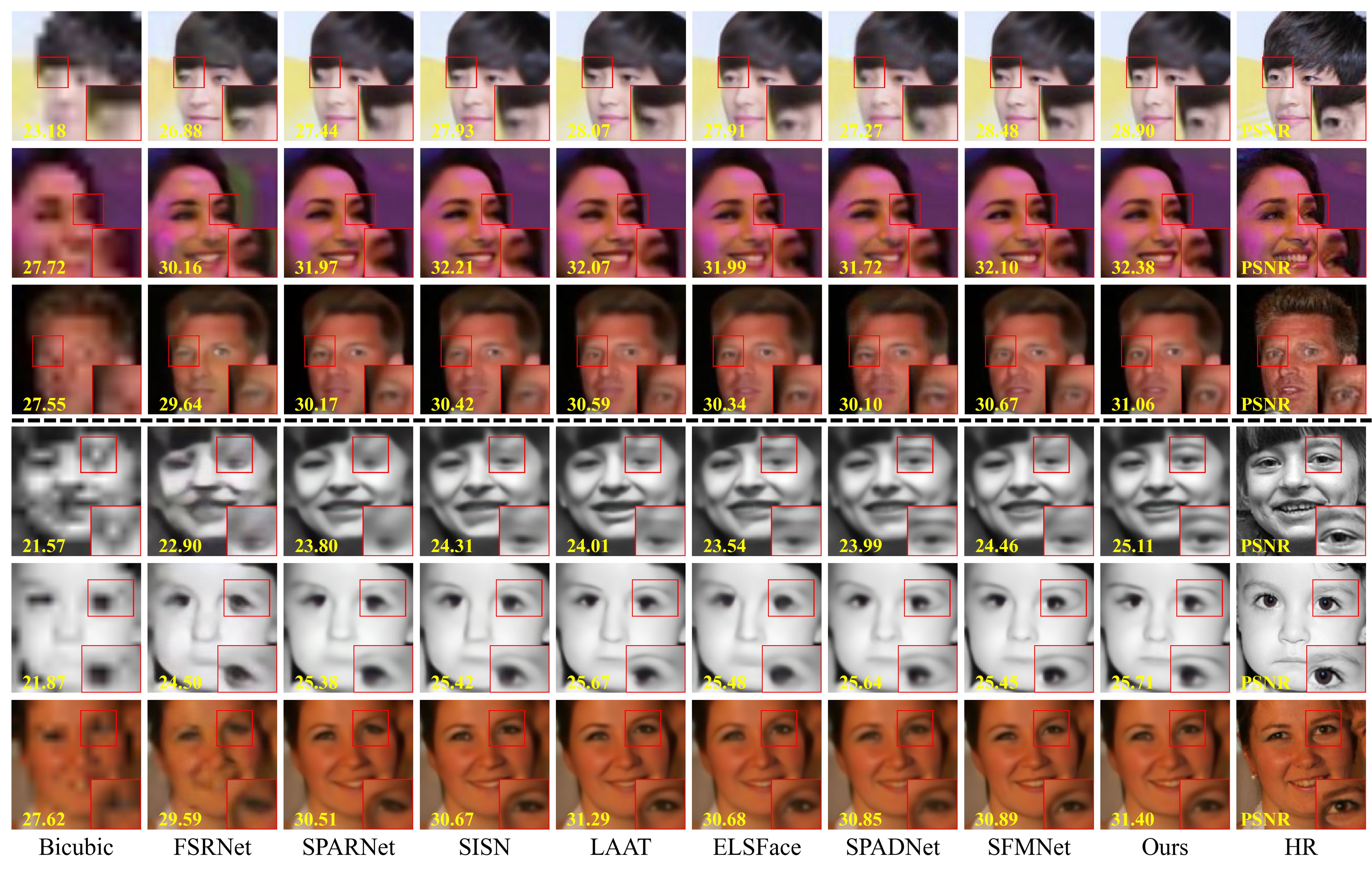}}
 \caption{Visual comparisons for $\times$8 FSR on CelebA test set~\cite{liu2015deep} and Helen test set~\cite{le2012interactive}. Please zoom in for more details.}
 \label{compare_CelebA}
\end{figure*}

\begin{figure}[t]
\centering
\includegraphics[width=1\columnwidth]{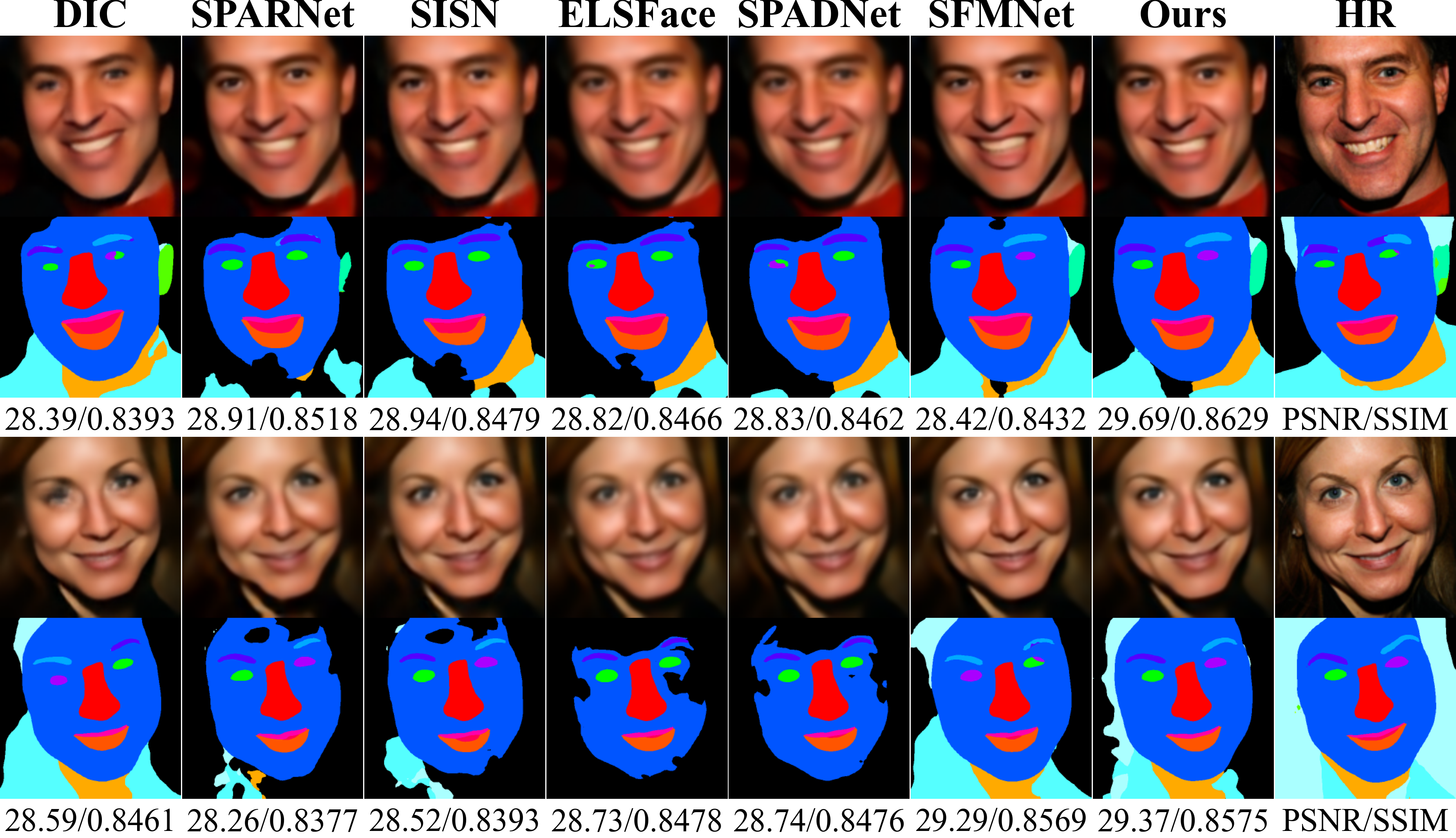}
\caption{Visual comparisons with existing methods on downstream tasks, like face parsing.}
\label{Face_parsing}
\end{figure}

\subsubsection{Study of local-global attention strategy} As summarized in Table~\ref{Frequency_ablation}, our goal is to examine the individual contributions of local and global modeling components in both high- and low-frequency enhancement branches. First, we remove the squeeze-and-excitation block from LFEB. The performance drops on CelebA and Helen. This suggests that incorporating squeeze-and-excitation operations to extract low-frequency global interactions and emphasize informative channels is beneficial, even in smooth regions. These global interactions help preserve structurally important facial components that might otherwise be weakened in low-frequency representations. Next, we remove the DPA module from HFEB. The result shows a more noticeable degradation, with performance dropping to 28.04 / 0.8032 on CelebA and 26.80 / 0.7981 on Helen. This highlights the crucial role of DPA in capturing localized textures and reducing artifacts in high-frequency content. Finally, removing the offset mechanism causes a moderate decline in performance, indicating that it contributes to cross-scale feature alignment.

To provide a more informative analysis of the offset mechanism, we further compare the proposed offset-based alignment module with several alternatives, including: (1) removing the alignment branch entirely, (2) replacing the learned offsets with a fixed warp, (3) using zero-offset warping, and (4) replacing it with a lightweight convolutional alignment branch. As shown in Table~\ref{tab:offset_ablation}, replacing the learned offset warp with a fixed warp, zero-offset warp, or convolution-based alignment fails to consistently mitigate cross-scale feature misalignment and leads to inferior overall performance across datasets and metrics. This indicates that the improvement does not simply come from adding an extra refinement branch or a generic warping operator, but specifically from the learned offset-based alignment itself.

\begin{figure}[t]
\centering
\includegraphics[width=1\columnwidth]{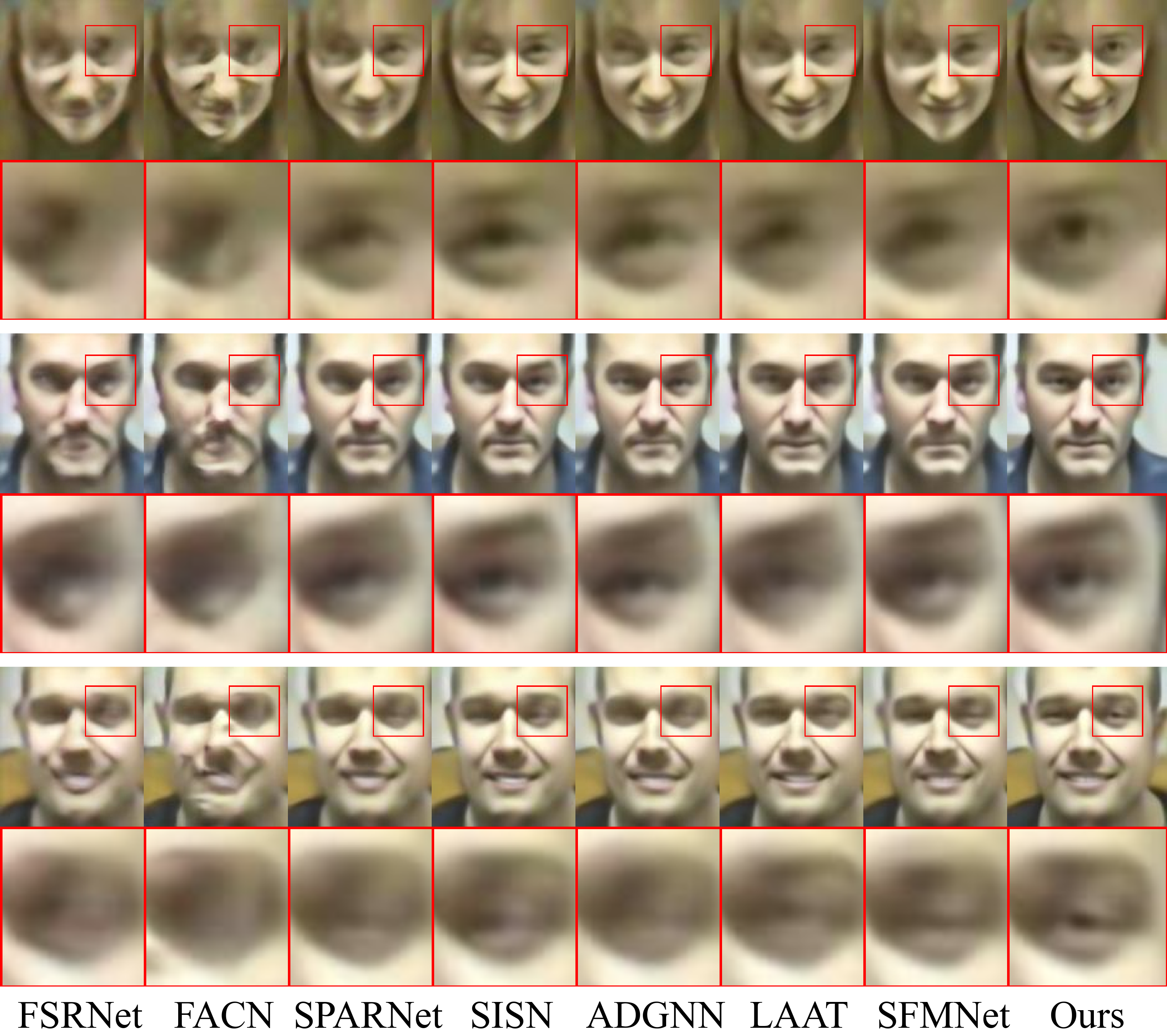}
\caption{Visual comparison on SCface~\cite{grgic2011scface} of real surveillance scenarios for $\times$8 SR. Our FADPNet achieves better FSR.}
\label{compare_SCface}
\end{figure}

\begin{table}[t] 
 \begin{center}
  \caption{Comparisons on face average similarity with existing methods on SCface~\cite{grgic2011scface} of real surveillance scenarios.}
  \setlength{\tabcolsep}{2.5mm}
        \renewcommand\arraystretch{1.1}
  \scalebox{1}{
  \begin{tabular}{l|c|c|c|c|c}
   \toprule
   \multirow{2}{*}{Methods} &   \multicolumn{5}{c}{Average Similarity}     \\
   &Case 1 &Case 2 &Case 3 &Case 4 &Case 5 \\
    \hline
    \hline
  RCAN~\cite{zhang2018image}      &0.5481 &0.5776 &0.4441 &0.7139 &0.5754  \\
  FSRNet~\cite{chen2018fsrnet}    &0.4930 &0.5212 &0.5487 &0.6544 &0.5276  \\
  FACN~\cite{xin2020facial}       &0.3670 &0.3309 &0.3326 &0.5612 &0.4373  \\
  SPARNet~\cite{chen2020learning} &0.6382 &0.6412 &0.6455 &0.6670 &0.6259  \\
  SISN~\cite{lu2021face}          &0.6182 &0.6471 &0.6354 &0.7011 &0.6469  \\
  AD-GNN~\cite{bao2022attention}  &0.6151 &0.6664 &0.6432 &0.7213 &0.6034  \\
  LAAT~\cite{li2023learning}      &0.6083 &0.7394 &0.7282 &0.7157 &0.5965  \\
  SFMNet~\cite{wang2023spatial}   &0.6240 &0.6048 &0.6330 &0.7158 &0.5742  \\
  \bf{FADPNet }                       &\bf{0.6782}  &\bf{0.7412} &\bf{0.7592} &\bf{0.7714} &\bf{0.6623}  \\
  \bottomrule
  \end{tabular}}
 \label{Average_Similarity}
 \end{center}
\end{table}

\begin{figure}[t]
\centering
\subfloat[PSNR-Speed-FLOPs]{
    \includegraphics[width=0.49\linewidth,trim=30 10 0 0]{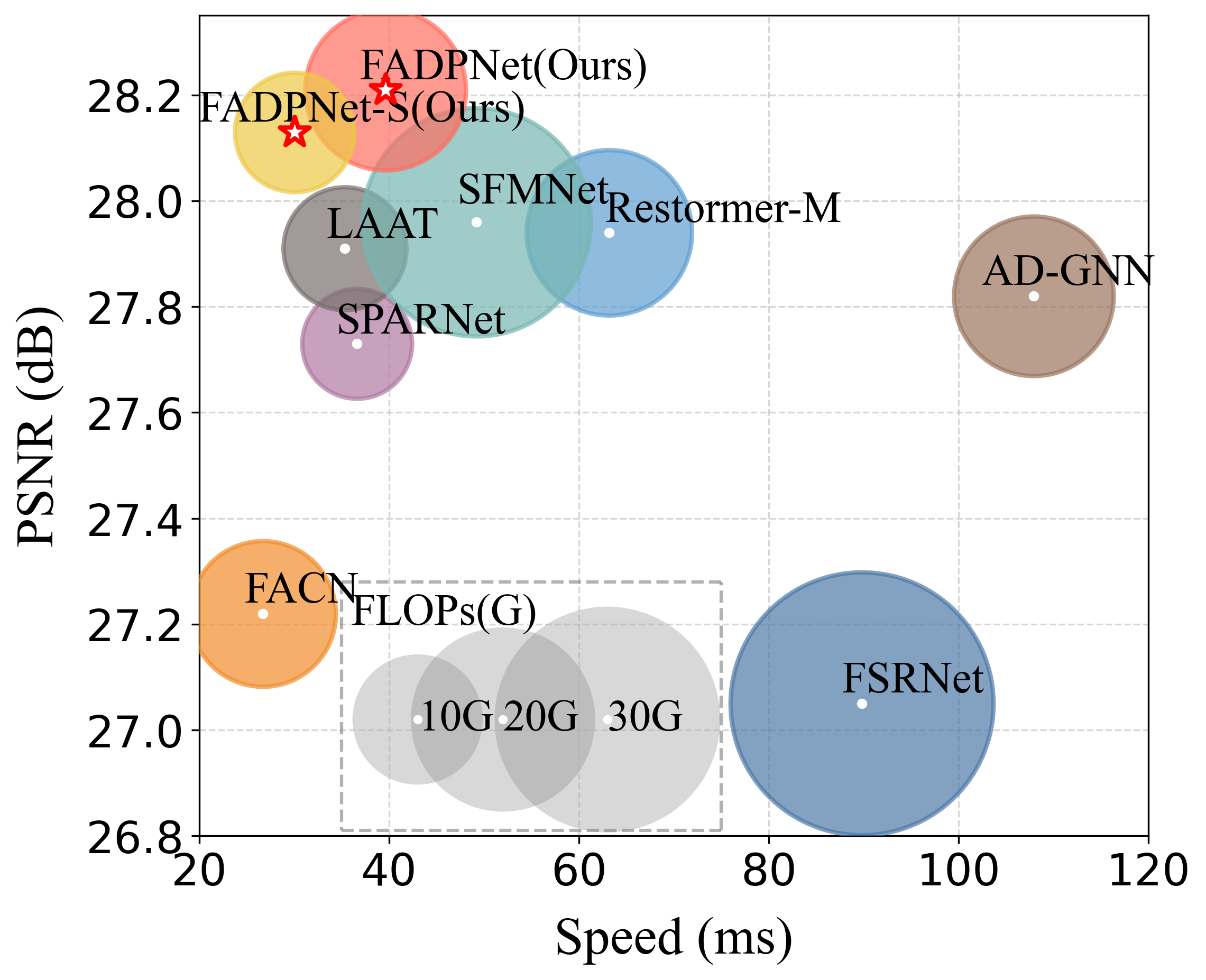}
    \label{flops_speed}}
\subfloat[PSNR-Params-FLOPs]{
    \includegraphics[width=0.48\linewidth,trim=30 15 0 0]{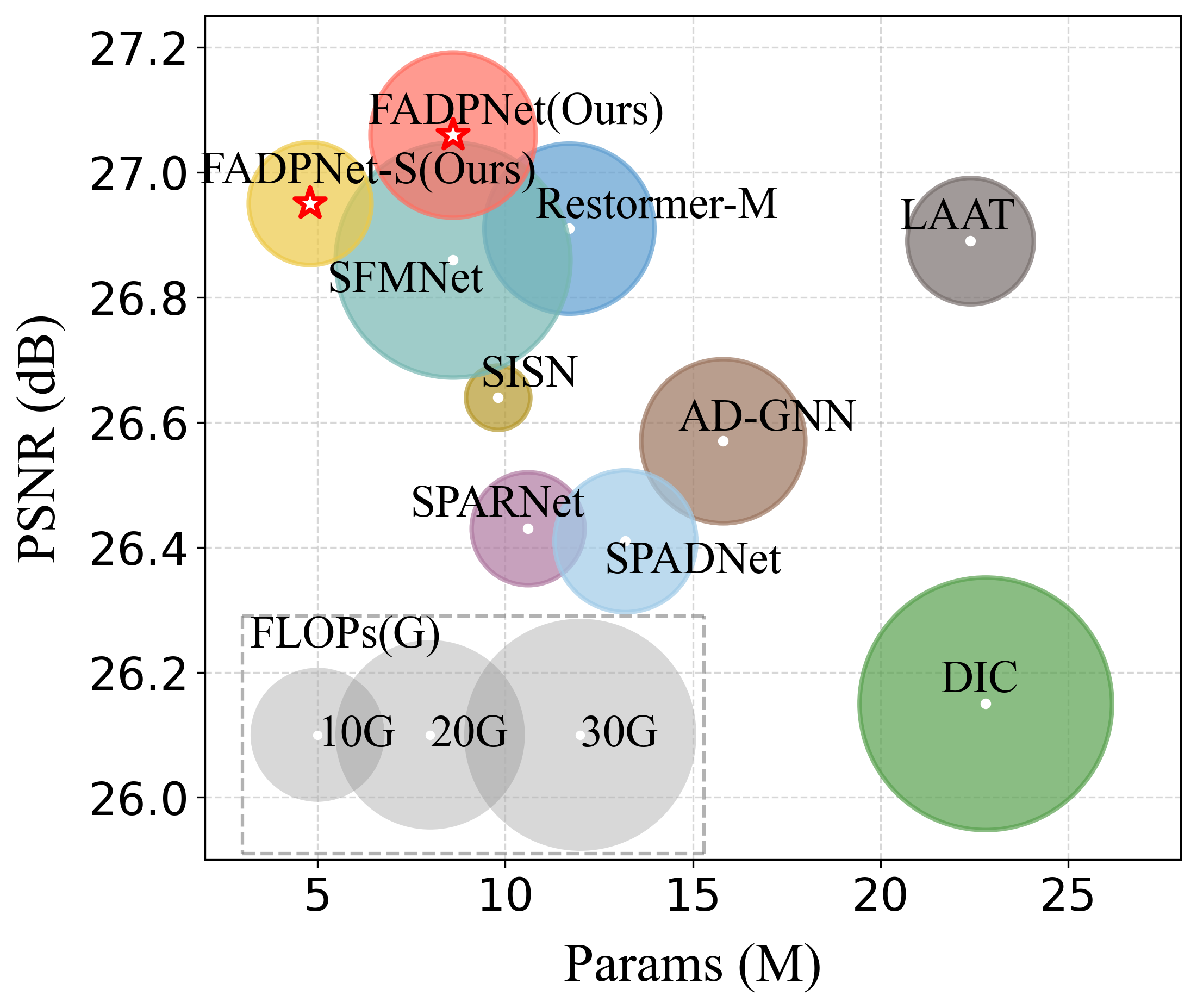}
    \label{flops_params}}
\caption{(a) PSNR, FLOPs, and speed tradeoffs on the CelebA test set~\cite{liu2015deep}. (b) PSNR, FLOPs, and Params tradeoffs on the Helen test set~\cite{le2012interactive}, showing the balance of our method.}
\label{flops}
\end{figure}

\subsection{Comparison with Other Methods}
\label{sec44}
We compare our method with existing FSR methods, encompassing general image SR methods like Restormer-M~\cite{zamir2022restormer}, FreqFormer~\cite{dai2024freqformer} and FreMamba~\cite{xiao2024frequency}, specialized FSR approaches such as FSRNet~\cite{chen2018fsrnet}, DIC~\cite{ma2020deep}, SPARNet~\cite{chen2020learning}, AD-GNN~\cite{bao2022attention}, SISN~\cite{lu2021face}, LAAT~\cite{li2023learning}, SFMNet~\cite{wang2023spatial}, ELSFace~\cite{qi2023efficient}, SPADNet~\cite{wang2024structure} and DMNet~\cite{li2026dual}. All methods are trained on the CelebA~\cite{liu2015deep} dataset under identical settings—including preprocessing, loss functions, and optimizer configurations—to eliminate implementation biases.

\subsubsection{Comparison on the CelebA dataset and the Helen dataset}
Table~\ref{compare_CelebA_Helen} reports quantitative results on CelebA and Helen, with the best and second-best scores marked in bold and underlined, respectively. Our model, trained on CelebA and directly applied to Helen, achieves state-of-the-art $\times$8 FSR performance, consistently balancing structural accuracy and perceptual quality. With only 8.6M parameters—similar to SFMNet~\cite{wang2023spatial}—FADPNet runs at 39.6 ms, 1.8× faster than FreqFormer~\cite{dai2024freqformer} (75 ms) and comparable to SPARNet~\cite{chen2020learning} (36.6 ms). Compared to ELSFace~\cite{qi2023efficient}, it improves PSNR by 0.25dB on CelebA and 0.36dB on Helen. Compared with those recent frequency-aware methods such as FreMamba~\cite{xiao2024frequency} and DMNet~\cite{li2026dual}, our FADPNet still maintains the best performance and faster inference speed on both CelebA and Helen datasets, further demonstrating the effectiveness of the proposed frequency-dispatched design. Our lighter variant, denoted as FADPNet-S (the channel width is reduced, while the overall frequency-aware dual-path framework is preserved), still achieves competitive performance with lower computational cost. As also shown in Fig.~\ref{compare_CelebA}, competing methods often yield blurred contours or distorted features (e.g., ill-defined eyes), whereas our reconstructions preserve sharp edges, facial symmetry, and fine details (e.g., eyeball), closely aligning with the ground-truth HR images. Face parsing results on Helen (as shown in Fig.~\ref{Face_parsing}) further confirm that FADPNet produces the most complete and accurate facial structures. Overall, our method outperforms prior methods in both performance (PSNR) and efficiency (Params-FLOPs-Speed) while maintaining robust generalization across datasets.

\subsubsection{Comparison on real-world surveillance}
Restoring facial details from real surveillance imagery is challenging due to LR, uncontrolled lighting, and sensor noise—conditions often underrepresented in synthetic benchmarks. We evaluate on SCface~\cite{grgic2011scface}, which contains naturally degraded surveillance images without manual downsampling. As shown in Fig.~\ref{compare_SCface}, existing methods frequently produce blurred contours and lose texture details under such conditions. In contrast, FADPNet, with adaptive feature separation and dual-frequency enhancement, recovers sharper structures, finer textures, and natural facial symmetry. We further assess downstream performance in face matching by comparing restored faces with high-definition references. Table \ref{Average_Similarity} shows that FADPNet consistently achieves higher similarity scores, confirming its practical effectiveness in challenging surveillance scenarios.


\subsubsection{Model Complexity Analysis}
Beyond FSR quality, we evaluate model efficiency in terms of parameters and inference speed, critical for resource-constrained devices. At each stage, inputs are decomposed into high- and low-frequency components, processed respectively by a CNN branch for high-frequency facial details and a Mamba-based branch for low-frequency, long-range dependencies. This design balances computational efficiency with strong representation of fine details. As shown in Fig.~\ref{flops_speed} and Fig.~\ref{flops_params}, FADPNet achieves superior PSNR with lower FLOPs, faster inference, and fewer parameters compared to existing methods, consistently outperforming them in both accuracy and model cost.

 \section{Limitations and Future Works}
Despite the dual-path design, the current fusion strategy is relatively simplistic and may not fully exploit the complementary nature of low- and high-frequency information, leading to suboptimal fine-grained detail recovery. In addition, the model shows a performance bias toward frontal faces, with reduced robustness to side views or large pose variations due to limited training view diversity. Future work will investigate more effective fusion mechanisms to enhance inter-frequency interaction, as well as pose-aware augmentation strategies to improve generalization in real-world applications.

\section{Conclusions}
In this paper, we propose FADPNet, a frequency-aware dual-path network that separates and processes low- and high-frequency components for the frequency imbalance problem in face super-resolution. The Low-Frequency Enhancement Block (LFEB), comprising ASSB and SEB, models global context and reinforces identity-consistent features. In parallel, the High-Frequency Enhancement Block (HFEB), built with DPA and HFR, refines local structures through adaptive attention mechanisms. This frequency-specific design enables more effective resource allocation and promotes both structural coherence and fine-detail recovery. Extensive experiments validate the superiority of our method in the reconstruction of face image quality and computational efficiency, underscoring the effectiveness of our proposed strategy.

\bibliographystyle{IEEEtran}
\bibliography{reference}

%

\begin{IEEEbiographynophoto}{Siyu Xu}
received the B.S. degree in Automation from the College of Intelligent Science and Control Engineering, Jinling Institute of Technology, Nanjing, China, in 2022, and the M.S. degree in lectronic Information from the College of Automation, Nanjing University of Posts and Telecommunications, Nanjing, China, in 2026. His research interest includes image super-resolution.
\end{IEEEbiographynophoto}

\begin{IEEEbiographynophoto}{Wenjie Li}
received the M.S. degree in control science and engineering from the College of Automation, Nanjing University of Posts and Telecommunications, Nanjing, in 2023. He is currently pursuing the Ph.D. degree in artificial intelligence with the School of Artificial Intelligence, Beijing University of Posts and Telecommunications. His research interests include image restoration.
\end{IEEEbiographynophoto}

\begin{IEEEbiographynophoto}{Guangwei Gao}
(Member, IEEE) received the Ph.D. degree in Pattern Recognition and Intelligent Systems from the Nanjing University of Science and Technology (NJUST), Nanjing, China, in 2014. He is currently a Professor with the School of Computer Science and Engineering, NJUST. He has served as an Associate Editor for \textsc{Pattern Recognition} and \textsc{IEEE Transactions on Image Processing}. His research interests include pattern recognition and computer vision. Personal website: \textit{https://guangweigao.github.io}.
\end{IEEEbiographynophoto}

\begin{IEEEbiographynophoto}{Jian Yang}
(Member, IEEE) received the Ph.D. degree in Pattern Recognition and Intelligent Systems from the Nanjing University of Science and Technology (NJUST), Nanjing, China, in 2002. 
He is currently a professor with the School of Computer Science and Engineering, NJUST. He is the author of more than 400 scientific papers in pattern recognition and computer vision. 
His research interests include pattern recognition, computer vision, and machine learning. He is/was an associate editor for \textsc{Pattern Recognition} and \textsc{IEEE Transactions on Neural Networks and Learning Systems}. He is a fellow of IAPR.
\end{IEEEbiographynophoto}

\begin{IEEEbiographynophoto}{Guo-Jun Qi} 
(Fellow, IEEE) has been a faculty member with the Department of Computer Science, University of Central Florida, since August 2014. He is currently a Professor and the Chief Scientist who oversees the Artificial Intelligence Research Center, Westlake University, and the OPPO U.S. Research Center. His research interests include machine learning and knowledge discovery from multi-modal data to build smart and reliable information and decision-making systems.
\end{IEEEbiographynophoto}

\begin{IEEEbiographynophoto}{Chia-Wen Lin}
(Fellow, IEEE) received the Ph.D. degree in Electrical Engineering from National Tsing Hua University (NTHU), Hsinchu, Taiwan, in 2000. 
He is currently a Distinguished Professor with the Department of Electrical Engineering and the Institute of Communications Engineering, NTHU. 
His research interests include image and video processing, computer vision, and video networking. Currently, he is serving as an Associate Editor-in-Chief of \textsc{IEEE Transactions on Circuits and Systems for Video Technology}.
\end{IEEEbiographynophoto}





\end{document}